%% file: main-R2.tex
\documentclass[5p,times,twocolumn]{elsarticle}

\usepackage{amssymb}
\usepackage{lineno,hyperref}
\usepackage{url}            
\usepackage{booktabs}       
\usepackage{amsfonts}       
\usepackage{nicefrac}       
\usepackage{microtype}      
\usepackage{xcolor}         
\usepackage{color, colortbl}
\usepackage{graphicx}
\usepackage{multirow}
\usepackage{svg}
\usepackage[bold]{hhtensor}
\usepackage{afterpage}
\usepackage{float}
\usepackage{placeins}
\newcommand{\etal}{\text{et al.}}
\usepackage[caption=false,position=bottom]{subfig}

\newcommand{\pmerror}[2]{#1\scriptsize{\ensuremath{\pm}}{\scriptsize #2}}
\usepackage{pifont}
\newcommand{\cmark}{\ding{51}}%
\newcommand{\xmark}{\ding{55}}%

\journal{Pattern Recognition}
\begin{document}

\begin{frontmatter}

\title{SoftPatch+: Fully Unsupervised Anomaly Classification and Segmentation}

\author[1address,2address]{Chengjie Wang\fnref{equalcontribution}} 
\author[3address]{Xi Jiang\fnref{equalcontribution}}               
\author[2address]{Bin-Bin Gao}            
\author[2address]{Zhenye Gan}             
\author[2address]{Yong Liu}               
\author[3address]{Feng Zheng}             
\author[1address]{Lizhuang Ma\corref{mycorrespondingauthor}}
\cortext[mycorrespondingauthor]{Corresponding author}
\fntext[equalcontribution]{Equally contribution.}
\ead{ma-lz@cs.sjtu.edu.cn}

\address[1address]{Department of Computer Science and Engineering, Shanghai Jiao Tong University, China}
\address[2address]{Youtu Lab, Tencent, China}
\address[3address]{Department of Computer Science and Engineering, Southern University of Science and Technology, China}

\input{secs/0_abstract.tex}

\end{frontmatter}

\input{secs/1_introduction.tex}
\input{secs/2_related_work.tex}
\input{secs/3_method.tex}
\input{secs/4_experiments.tex}

\input{secs/5_conclusion.tex}

\bibliography{sp}

\end{document}

%% file: secs/0_abstract.tex
\begin{abstract} \label{section:abs}
Although mainstream unsupervised anomaly detection (AD) (including image-level classification and pixel-level segmentation) algorithms perform well in academic datasets, their performance is limited in practical application due to the ideal experimental setting of clean training data. Training with noisy data is an inevitable problem in real-world anomaly detection but is seldom discussed. This paper {is the first to consider} fully unsupervised {industrial} anomaly detection (i.e., unsupervised AD with noisy data). To solve this problem, we proposed memory-based unsupervised AD methods, SoftPatch and SoftPatch+, which efficiently denoise the data at the patch level.  
Noise discriminators are utilized to generate outlier scores for patch-level noise elimination before coreset construction. The scores are then stored in the memory bank to soften the anomaly detection boundary. 
Compared with existing methods, SoftPatch maintains a strong modeling ability of normal data and alleviates the overconfidence problem in coreset, and SoftPatch+ has more robust performance which is particularly useful in real-world industrial inspection scenarios with high levels of noise (from 10\% to 40\%).
Comprehensive experiments {conducted in diverse noise scenarios} demonstrate that both SoftPatch and SoftPatch+ outperform the state-of-the-art AD methods on the MVTecAD, ViSA, and BTAD benchmarks. {Furthermore, the performance of SoftPatch and SoftPatch+ is} comparable to {that of the noise-free methods in conventional unsupervised AD setting. The code of the proposed methods can be found at https://github.com/TencentYoutuResearch/AnomalyDetection-SoftPatch.}
\end{abstract}

%% file: secs/1_introduction.tex
\section{Introduction} \label{section:intro}
Detecting anomalies by only nominal images without annotation is an appealing topic, especially in industrial applications where defects can be extremely tiny and hard to collect.
Recent deep learning methods~\citep{dsr,ding2022catching} usually model the AD problem as a one-class learning problem and employ computer visual tricks to improve the perception where a clean nominal training set is provided to extract representative features. 
{Most previous unsupervised AD methods rely on a clean training dataset to model a standard distribution, which will be used to determine whether a test sample is normal or abnormal.}
Even though recent methods have achieved excellent performance, they all rely on the clean training set to extract nominal features for later comparison with anomalous features. 
Putting too much faith in training data can lead to pitfalls. 
If the standard normal dataset is polluted with noisy data, i.e., the defective samples, the estimated boundary will be unreliable, and the classification for abnormal data will have low accuracy. In general, unsupervised AD methods are not designed for and are not robust to noisy data. {From the perspective of practical applications, it is challenging to ensure that the data from the production line is completely clean~(i.e., zero noise). Therefore, it is crucial to study unsupervised AD with noise.}

{Learning from noisy data has been a long-standing challenge in traditional machine learning and modern deep learning. Deep learning models typically require more training data than machine learning models. In many applications, the training data are labeled by non-experts. Therefore, the noise ratio may be higher in these datasets compared with the smaller
and more carefully prepared datasets used in classical machine
learning. 
Many studies have demonstrated the positive impact on the performance of models when the noise is eliminated or denoised to a certain extent. 
For image classification, simPLE \citep{simPLE}, FixMatch\citep{fixmatch}, and Dividemix \citep{li2020dividemix} are proposed to filter noisy pseudo-labeled data. For object detection, multi-augmentation~\cite{softteacher}, teacher-student~\cite{unbiased}, or contrastive learning~\cite{CCSSL} are adopted to alleviate noise with the help of the expert model's knowledge. However, these methods rely on labeled data to co-rectify noisy data. In comparison, we aim to improve the robustness in an unsupervised manner without introducing labor annotations,  which have rarely been explored in unsupervised AD.}

\begin{figure*}
  \centering
  \includegraphics[width=1.0\linewidth]{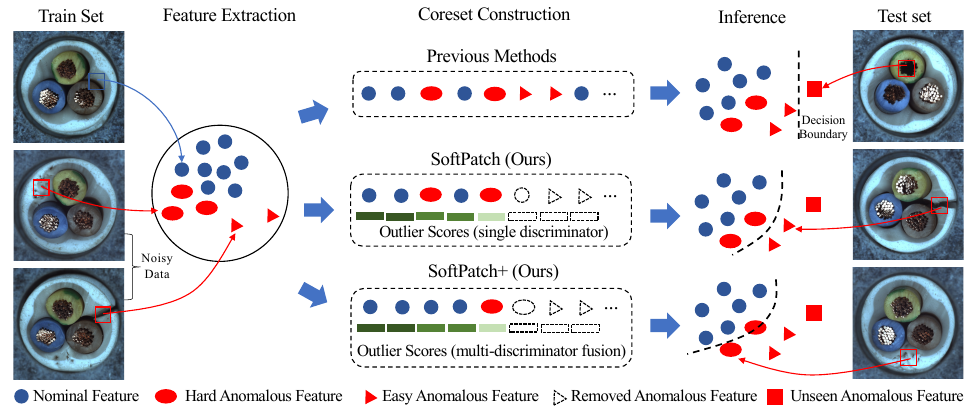}
  \caption{Illustration of SoftPatch/SoftPatch+. Unlike previous methods that construct coreset without considering the negative effect of noisy data, SoftPatch identifies and excludes outliers using an outlier score. When there is excessive noise, a single outlier discriminator's capability is limited. So, SoftPatch+ further improves robustness by using the fusion of multiple discriminators.}
  \label{fig:1}
\end{figure*}

In real-world practice, it is inevitable that there are noises that sneak into the standard normal dataset, especially for industrial manufacturing, where a large number of products are produced daily. The ratio of corrupted products may reach 38.5\% in some real-world applications as reported in \citep{song2022learning}.
{
Suppose the ratio of good production is greater than 60\%, meaning the noise rate is less than 40\%. In that case, the raw data from the production line may contain up to 40\% noise, and it is necessary to develop unsupervised AD algorithms for handling noisy training data. On the other hand, when the ratio of good production is below 60\%, efforts are generally focused on optimizing and improving the good production rate rather than unsupervised AD algorithms.
}
This noise usually comes from the inherent data shift or human misjudgment. 
Meanwhile, existing unsupervised AD methods, {such as PatchCore~\cite{roth2021towards}, CFLOW~\cite{gudovskiy2022cflow}, and C2FAD~\cite{zheng2022focus}}, are susceptible to noisy data due to their exhaustive strategy to model the training set. 
As in Figure~\ref{fig:1}, noisy samples easily misinform those overconfident AD algorithms, so algorithms misclassify similar anomaly samples in the test set and generate wrong locations. 
Additionally, AD with noisy data can be developed to a fully unsupervised setting, which discards the implicit supervised signal that the training set is all defect-free, compared with the previous unsupervised setting in AD. 
This setting helps to expand more industrial quality inspection scenarios, i.e., rapid deployment to new production lines without data filtration. 

In this paper, we first point out the significance of studying fully unsupervised {industrial} AD. 
Our solution is inspired by one of the recent state-of-the-art methods, PatchCore~\cite{roth2021towards}. PatchCore proposed a method to subsample the original CNN features of the standard normal dataset with the nearest searching and establish a smaller coreset as a memory bank. However, the coreset selection and classification process are vulnerable to polluted data. 
In this regard, we propose a patch-level selection strategy to wipe off the noisy image patch of noisy samples. {Compared to conventional image-level denoising, the abnormal patches are discarded, and the normal patches of a noise sample are still be used to construct coreset.} 
Specifically, the denoising algorithm assigns an outlier factor to each patch to be selected into the coreset. Based on the patch-level denoising, we propose a novel AD algorithm with better noise robustness named SoftPatch. 
Considering noisy samples are hard to remove completely, SoftPatch utilizes the outlier factor to re-weight the coreset examples. Patch-level denoising and re-weighting the coreset samples are proved effective in revising misaligned knowledge and alleviating the overconfidence of coreset in inference. 
{Extensive experiments in various noise scenes demonstrate that SoftPatch and SoftPatch+ outperforms the state-of-the-art (SOTA) AD methods on MVTec Anomaly Detection (MVTecAD) \cite{16}, VisA \cite{visa} and BTAD~\cite{btad} benchmark. }

Our main contributions are summarized as follows:
\begin{itemize}
    \item[$\bullet$] 
To the best of our knowledge, we are the first to focus on fully unsupervised {industrial} anomaly detection, which is a more practical setting but seldom investigated. 
Existing unsupervised AD methods fully trust the training set's cleanliness, leading to their performance degradation in noise interference. 
    \item[$\bullet$] 
We propose a patch-level denoising strategy for the coreset memory bank, which essentially improves the data usage rate compared to conventional sample-level denoising. Based on this strategy, we apply three discriminators which strengthen model robustness by combining the re-weighting of coreset.
    \item[$\bullet$] 
We set a baseline for fully unsupervised {industrial} AD, which performs well in the settings with additional noisy data and the general settings without noise, providing a new view for further research. 

\end{itemize}

The preliminary version of this work (SoftPatch \cite{softpatch}) was presented {at the conference NeurIPS 2022}. We extend the original work in several significant ways in this journal version. 
{\textbf{\textit{First}}, we comprehensively build no-overlap and overlap fully unsupervised AD settings to make it closer to real-world industrial inspection scenarios increasing the noise ratio from 10\% to 40\% empirically~(\textbf{Sec. 4}). \textbf{\textit{Second}}, we propose a new method called SoftPatch+ by multi-discriminator (\textbf{Sec. 3.3}), {which enhances the robustness and achieves better performance compared to existing methods}, especially in scenarios with higher noise ratios.
\textbf{\textit{Finally}}, extensive experiments are conducted on more datasets, like VisA~\cite{visa}, to demonstrate the effectiveness of SoftPatch and SoftPatch+ (\textbf{Secs. 5.2} and \textbf{5.3}). More comprehensive visualization and analysis are provided to fully demonstrate the effectiveness of our approach.}

%% file: secs/2_related_work.tex
\section{Related Work} \label{section:related}

\subsection{Unsupervised Anomaly Detection}

\textbf{Training with {pretext} tasks.} Also known as self-supervised learning, {pretext} tasks are a viable solution when there is no category and shape information of anomalies. 
Sheynin~\etal ~\cite{5} employs transformations such as horizontal flip, shift, rotation, and gray-scale change after a multi-scale generative model to enhance the representation learning. Li~\etal ~\cite{2} mentions that naively applying existing self-supervised tasks is sub-optimal for detecting local defects and {proposes} a novelty {pretext} task named CutPaste, which simulates an abnormal sample by clipping a patch of a standard image and pasting it back at a random location. 
Similarly, DRAEM~\citep{7} synthesizes anomalies through Perlin Noise. {Recently, SaliencyCut~\cite{ye2024saliencycut} proposes a saliency-guided data augmentation for creating pseudo anomalies.}
Nevertheless, the inevitable discrepancy between the synthetic anomaly and the real anomaly disturbs the criteria of the model and limits the generalization performance. The gap between anomalies is usually larger than that between anomaly and normal. This is why AD methods deceived by some noisy samples can still work well when handling other kinds of anomalies.   

\textbf{Agnostic methods.} Including knowledge distillation and image reconstruction, agnostic methods based on a theory that models that have never seen anomalies will behave differently in inference when inputting both normal and anomaly samples. 
Knowledge distillation is ingeniously used in anomaly detection. Bergmann~\etal~\cite{11} proposes that the representations of unusual patches are different between a pre-trained teacher model and a student model, which tried its best to simulate teacher output with an anomaly-free training set. Based on this theory,
Salehi~\etal~\cite{4} proposes that considering multiple intermediate outputs in distillation and using a smaller student network {leads} to a better result. 
{Building upon the density-based landmark selection, AADAE~\cite{zhu2022adaptive} uses the adaptive aggregation-distillation autoencoder for unsupervised anomaly detection.}
Reverse distillation~\citep{deng2022anomaly} uses a reverse flow that avoids the confusion caused by the same filters and prevents the propagation of anomaly perturbation to the student model, whose structure is similar to reconstruction networks.
{Similar to the idea of reverse distillation, DFC~\cite{yang2022learning} develops an asymmetric dual framework that consists of a generic feature extraction network and an elaborated feature estimation network, and detects the possible anomalies within images by modeling and evaluating the associated deep feature correspondence between the two dual network branches.}
Image Reconstruction methods~\cite{zavrtanik2021reconstruction,liang2022omni} {utilizes} the assumption that the reconstruction network trained in the normal set can not reconstruct the anomaly part. A high-resolution result can be obtained by comparing the differences between the reconstructed and original images. {Recently, a novel gating highway connection module has been proposed to adaptively integrate skip connections into the reconstruction-based framework and boost its anomaly detection performance, namely GatingAno~\citep {zhang2024anomaly}.}
However, all agnostic methods need long training stages, which limit their usage, i.e., the rapid deployment assumption in fully unsupervised learning.

\textbf{Feature modeling.} We specifically refer to the direct modeling of the output features of the extractor, including distribution estimation~\citep{defard2021padim}, distribution transformation~\citep{9,gudovskiy2022cflow}, graph model-based multiscale feature\cite{zhang2023graph}, pre-trained model adaption~\citep{yang2022learning,lee2022cfa}, and memory storage~\citep{6, roth2021towards}. 
PaDiM~\citep{defard2021padim} utilizes multivariate Gaussian distributions to estimate the patch embedding of nominal data. In the inference stage, the embedding of irregular patches will be out of distribution. It is a simple but efficient method, but Gaussian distribution is inadequate for more complex data cases. 
So to enhance the estimation of density, DifferNet~\citep{9} and CFLOW~\citep{gudovskiy2022cflow} leverage the reversible normalizing flows based on multi-scale representation. 
Hou\etal~\cite{6} proposed that the granularity of division on feature maps is closely related to the reconstruction capability of the model for both normal and abnormal samples. So a multi-scale block-wise memory bank is embedded into an autoencoder network as a model of past data. 
PatchCore~\citep{roth2021towards} is a more explicit but valuable memory-based method, which stores the sub-sampled patch features in the memory bank and calculates the nearest neighbor distance between the test feature and the coreset as an anomaly score. {Similar to PatchCore, Graph-Construction~\cite{zhang2023graph} uses a multiscaled graph model to establish the relation between query and memory bank in the nearest neighbor distance.}
Although PatchCore and Graph-Construction are outperformance in the typical setting, they are overconfident in the training set, which leads to poor noise robustness.

\subsection{Learning with Noisy Data}

{Learning from noisy data has been a long-standing challenge in traditional machine learning and modern deep learning. Deep learning models typically require more training data than machine learning models. In many applications, the training data are labeled by non-experts. Therefore, the noise ratio may be higher in these datasets compared with the smaller
and more carefully prepared datasets used in classical machine
learning. Many studies have demonstrated the positive impact on the performance of models when the noise is eliminated or denoised to a certain extent. Different from these works, we aim to improve the robustness in an unsupervised manner without introducing labor annotations, which have rarely been explored in unsupervised AD. However, many existing denoising works rely on labeled data to co-rectify noisy data.
For image classification, simPLE \citep{simPLE}, FixMatch\citep{fixmatch}, and Dividemix \citep{li2020dividemix} are proposed to filter noisy pseudo-labeled data. For object detection, multi-augmentation~\cite{softteacher}, teacher-student~\cite{unbiased}, or contrastive learning~\cite{CCSSL} are adopted to alleviate noise with the help of the expert model's knowledge.}

\begin{figure*}[h]
  \centering
  \includegraphics[width=1.0\linewidth]{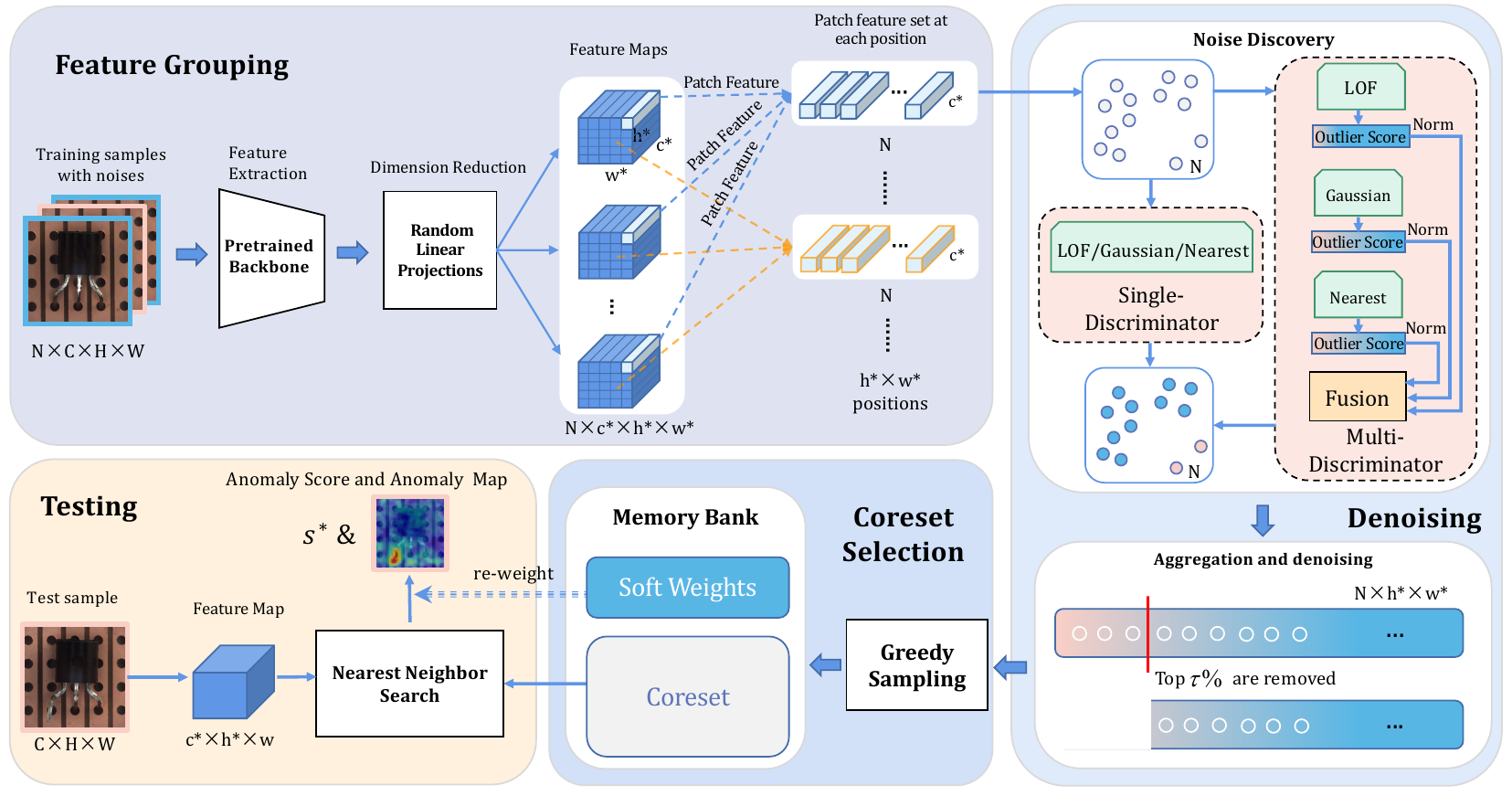}
  \caption{Overview of the proposed method. In the training phase, the noises are distinguished at patch-level at each position of the feature map by a single discriminator (SoftPatch) or multiple discriminators (SoftPatch+). The deeper color a patch node has, the higher probability that it is a noise patch. After achieving outlier scores for all patches, the top $\tau$\% patches with the highest outlier score are removed. The coreset is a subset of remaining patches after denoising. Different from other methods, our memory bank consists of the samples in coreset and their outlier scores which are stored as soft weights. Soft weights will be further utilized to re-weight the anomaly score in inference.}
  \label{fig:framwork}
\end{figure*}

While there are some model robustness researches~\cite{han2022adbench} on unsupervised AD, their {objectives} and tasks are distinguished from our work. More importantly, the methods in these studies can not be applied directly to solve fully unsupervised AD problems. 
Pang~\etal 
~\cite{pang2020self} deals with video anomaly without manually labeled data where information in consecutive frames can be exploited, {while} our work tackles anomaly detection from images. Other works, like RCA~\cite{liu2021rca}, eliminate noisy and corrupted data in semantic anomaly detection. Unlike semantic anomaly detection, we focus on image sensory anomaly detection~\cite{yang2024generalized}, which has recently raised much concern and contains a new task, anomaly localization. 
Recently, Chen~\etal~proposes {Interpolated Gaussian Descriptor (IGD)} method~\cite{chen2022deep} for unsupervised AD with contaminated data. The adversarial
interpolation in IGD is enforced to consistently learn a smooth Gaussian descriptor, even when the training data is contaminated with anomalous samples. Instead of IGD modeling on noisy data, our method aims to remove patch-level noisy samples using noise discriminators and then build an AD model based on the remaining clean samples.

%% file: secs/3_method.tex
\section{Methodology}\label{section:method}

\subsection{Overview}

Patch-based unsupervised anomaly methods, such as PatchCore \cite{roth2021towards} and CFA ~\cite{lee2022cfa},  have three main processes: feature extraction, coreset selection with memory bank construction, and anomaly detection. 
One of the important assumptions is that the training set only contains nominal images, and the coreset should have full coverage of the entire training data distribution. During the test, an incoming image will directly search in the memory bank for similar features, and the anomaly score is the dissimilarity with the nearest {patch}. 
 The searching process may collapse if the assumed clean full coverage memory bank contains noise. Therefore, we propose SoftPatch {and SoftPatch+}, which filters noisy data by a single discriminator before coreset construction and softens the searching process for down-weighting the hard unfiltered noisy samples.

The general denoising methods against label contamination at the {image level} are sub-optimal in image sensory anomaly detection.
The abnormalities in image sensory AD, represented by industrial defect detection and medical image analysis, usually occupy only a tiny area of the image. {At the image level, noisy data is hard to distinguish, but the local region's inherent deviation may be more remarkable in the patch level}. 
{As illustrated in Figure~\ref{fig:framwork},} we propose a patch-level denoising {framework~(SoftPatch/SoftPach+)} that works on the feature space to judge the noisy patch better. 
First, the collected features are grouped according to position to narrow the domain of noise discrimination. 
Then we insert the patch-level noise discrimination process before the coreset sampling, which generates the noise score according to the feature distribution of each position. 
Since most areas of the noisy image are usually anomaly-free, we remove those noisy patches and retain the rest to maximize the use of data.
At the same time, the rest denoising scores reflecting the behavior of clustering are used to scale the anomaly score in inference. The other parts of the algorithm, such as feature extraction, dimension reduction, coreset sampling, and nearest neighbor search, follow the baseline PatchCore\cite{roth2021towards}. 

{\textbf{Problem Formulation:}} The target of image-level denoising is to find $\mathcal{X}_{noise}$ from $\mathcal{X}$, where $\mathcal{X} = \{x_i:i \in[1,N], x_i \in \mathbb{R}^{C \times H \times W} \}$ denotes training images (channels $C$, height $H$, width $W$). 
Following convention in existing work\cite{roth2021towards}, we use $\phi_{i} \in \mathbb{R}^{c^* \times h^* \times w^*} $ as the feature map (channels $c^*$, height $h^*$, width $w^*$) of image $x_i \in \mathcal{X}$, $\phi_{i}(h, w) \in \mathbb{R}^{c^*}$ as the patch at $(h, w)$ on the aggregated feature map with dimension $c$. {The patch-level denoising aims to find $\mathcal{X}_{noise}= \{\phi_{i}(h, w)\} $ from $\mathcal{X}$.}

\subsection{{Noise Discovery via Single-Discriminator}}
With increasing training images, the memory can become exceedingly large and infeasible to discriminate noise by overall statistics.  
Therefore, we group all features by position and count their outlier scores. Then all the scores are aggregated to determine noise patches, after which we just remove the features with top $\tau$\ percent scores. 
{We also fuse multiple noise reduction methods and construct a more robust coreset for fully unsupervised anomaly detection in some scenarios of high noise ratios.}

\subsubsection{Nearest Neighbor}  \label{sec:near_neigh}
With the assumption that the amount of noisy samples $\mathcal X_{noise}$ is much less than clean samples $\mathcal X_{nominal}$, we set {nearest neighbor distance~\cite{nearest_wiki}} as our baseline where a large distance means an outlier. 
Given a set of images, $\phi \in \mathbb{R}^{N \times c^* \times h^* \times w^*}$ represents all features. Each patch's nearest neighbor distance $\mathcal{W}_i^{nn}$ is defined as:
\begin{equation}
    \mathcal{W}^{nn}_i(h, w) = \min_{n \in [1,N] , n \neq i }\|  \vec {\phi}_i(h, w) - \vec {\phi}_n(h, w)\|_2 \label{eq:nn_distance},
\end{equation}
We first calculate the distances, then take the minimum among batch dimensions (neighbor) as {$\mathcal{W}_i^{nn}$}. 
This method can discriminate apparent outliers but suffers from uneven distribution of different clusters, where some clusters can have large inter-distances and lead to being mistakenly threshed as noisy data. To treat all clusters equally, we propose another multi-variate Gaussian method to calculate the outlier score without the interference of different clusters' densities. 

\subsubsection{Multi-Variate Gaussian} \label{sec:multi_vgau}
With Gaussian's normalizing effect, all clean images' features can be treated equally. To apply Gaussian distribution on image characteristics dynamically, we calculate the inlier probabilities on the batch dimension for each patch $\phi_i(h, w)$, similar to {Sec.}~\ref{sec:near_neigh}. The {multivariate} Gaussian distribution {$\mathcal{N}(\mu_{h,w}, \Sigma_{h, w})$} can be formulated that $\mu_{h,w}$ is the batch mean of $\phi_i(h, w)$ and sample covariance ${\Sigma}_{h, w}$ is:
\begin{equation}
    {{\Sigma}_{h, w}} = \frac{1}{N - 1}\sum_{n = 1}^N{(\phi_n(h, w) - \mu_{h, w})(\phi_n(h, w) - \mu_{h, w})^T + \epsilon I} \label{eq:gau_dist},
\end{equation}
where the regularization term $\epsilon I$ makes ${\sum}_{h, w}$ full rank and invertible \cite{defard2021padim}. Finally, with the estimated multi-variate Gaussian distribution $\mathcal{N}(\mu_{h,w}, \sum_{h, w})$. 
Notice that we do not exclude the $x_i$ when calculating the $\mathcal{N}$, so all samples can share a joint distribution, which is much less computationally intensive than doing the calculations one by one. 
Mahalanobis distance is calculated as the noisy magnitude $\mathcal{W}_i^{mvg}(h, w)$ of each patch:
\begin{equation}
    \mathcal{W}_i^{mvg}(h, w) = \sqrt{(\phi_i(h, w) - \mu_{(h, w)})^T{\Sigma}_{h, w}^{-1}(\phi_i(h, w) - \mu_{(h, w)})} \label{eq:gau_weight} .
\end{equation}
A high Mahalanobis distance means a high outlier score. Even though Gaussian distribution normalizes and captures the essence of image characteristics, small feature clusters may be overwhelmed by large feature clusters. In the scenario of a prominent feature cluster and a small cluster in a batch, the small cluster may be out of 1-, 2- or 3-$\Sigma$ of calculated $\mathbb{N}(\mu_{h,w}, \Sigma_{h, w})$ and erroneously classified as outliers. After analyzing the above two methods, we need a method that can: 1. treat all image characteristics equally; 2. treat large and small clusters equally; 3. make high dimension calculation applicable. 

\subsubsection{Local Outlier Factor (LOF)} \label{sec:lof}
LOF\cite{LOF} is a local-density-based outlier detector used mainly on E-commerce for criminal activity detection. Inspired by LOF, we can solve the above-mentioned three questions in {Sec.}~\ref{sec:multi_vgau}: 1. Calculating the relative density of each cluster can normalize different density clusters; 2. Using local k-distance as a metric to alleviate the overwhelming effect of large clusters; 3. Modeling distance as normalized feature distance can be used on high-dimensional patch features.
Therefore, the k-distance-based absolute local reachability density $lrd_i(h, w)$ is first calculated as:
\begin{equation}
    lrd_i(h,w) = 1/(\frac{\sum_{b \in \mathcal{N}_k(\phi_i(h,w))} dist^{reach}_k(\phi_i(h,w), \phi_b(h,w))}{|\mathcal{N}_k(\phi_i(h,w))|}),
\end{equation}
\begin{equation}
    dist^{reach}_k(\phi_i(h,w), \phi_b(h,w)) = max(dist_k(\phi_b(h,w)), d(\phi_i(h,w), \phi_b(h,w))),
\end{equation}
where $d(\phi_i(h,w), \phi_b(h,w))$ is L2-norm, $dist_k(\phi_i(h,w))$ is the distance of kth-neighbor, $\mathcal{N}_k(\phi_i(h,w))$ is the set of k-nearest neighbors of $\phi_i(h,w)$ and $|\mathcal{N}_k(\phi_i(h,w))|$ is the number of the set which usually equal k when without repeated neighbors. 
With the local reachability density of each patch, the overwhelming effect of large clusters is largely reduced. To normalize local density to relative density for treating all clusters equally, the relative density $\mathcal{W}_i^{LOF}$ of image $i$ is defined below:
\begin{equation}
    \mathcal{W}^{LOF}_i(h, w) = \frac{\sum_{b \in \mathcal{N}_k(\phi_i(h,w))} lrd_b((h, w))}{|\mathcal{N}_k(\phi_i(h,w))| \cdot lrd_i(h,w)} .
\end{equation}
$\mathcal{W}^{LOF}_i(h, w)$ is the relative density of the neighbors over {the} patch's own, and represents a patch's confidence of inlier. 
Our experiments found that all three noise reduction methods above are helpful in data pre-selection before coreset construction, while $LOF$ provides the best performance. 
However, after visualization of our cleaned training set, we found that hard noisy samples, which are similar to nominal samples, are still hidden in the dataset. To further alleviate the effect of noisy data, we propose a soft re-weighting method that can down-weight noisy samples according to outlier scores. 

\subsection{Robust Noise Discovery via Multi-Discriminator}\label{sec:mdf}
As a local-density-based outlier detector, LOF can achieve high accuracy in a relatively low noise ratio. However, there is a risk of failure in the local-density mechanism (based on clusters and local k-distance) used in LOF when the noise ratio is high. On the contrary, the Gaussian method measures outliers through a probability distribution. Although it may have lower accuracy than LOF in low-noise scenarios, it exhibits stronger robustness in high-noise ratios. Based on the complementarity between LOF and Multi-Variate Gaussian, we integrated them to achieve a more comprehensive and effective approach.

We normalized the outlier values generated by the aforementioned methods based on ranking and then took the average as the output of the combined method. The normalization by rank is defined as:
\begin{equation}
\mathcal{W}^{{\prime}{m}}_{i}(h,w)=\frac{Rank(\mathcal{W}_{i}^{m}(h,w))}{|\mathcal{W}^{m}_{i}(h,w)|}, m \in \{nn, mvg, lof\},
\end{equation}
here $Rank(*)$ refers to the ordinal position of a value after sorting it in ascending order based on its numerical value. $|\mathcal{W}^{x}_{i}(h,w)|$ is the number of $\mathcal{W}^{x}_{i}(h,w)$ values. And the fused average distance can be calculated as the following equation:
\begin{equation}
    \mathcal{W}^{vote} = \frac{\sum s^{m} \cdot \mathcal{W}^{{\prime}^{m}}_{i}(h,w)}{\sum s^{m}}, s^{m} \in \{0,1\}.
\end{equation}

Considering the different sensitivity to noise between detection and segmentation tasks, we select a specific coreset for each task by removing top $\tau_{cls}$ and top $\tau_{seg}$ percent outlier score, respectively. Here $\tau_{seg}$ is apparently larger than $\tau_{cls}$ since the anomaly segmentation task is much more sensitive {than} the classification. That is to say, we build a more compact coreset for segmentation tasks.

\subsection{Anomaly Detection based on SoftPatch and SoftPatch+}
Besides the construction of the Coreset, outlier factors of all the selected patches are stored as soft weights in the memory bank. With the denoised patch-level memory bank $\mathcal{M}$ as shown in {Figure}~\ref{fig:framwork}, the image-level anomaly score $s \in \mathbb{R}$ can be calculated for a test sample $x_i \in \mathcal{X}^{test}$ by nearest neighbor searching at patch level. Denoting the collection of patch features of a test sample as {$\mathcal{P}_{x_i}$}, for each patch $p_{h,w} \in \mathcal{P}_{x_i}$ the nearest neighbor searching can be formulated as the following equation:
\begin{equation}
    m^* = \mathop{\arg\min}\limits_{m \in \mathcal{M}} \|p-m\|_2 .
\end{equation}

After nearest searching, pairs of test {patches} and their corresponding nearest neighbor in $\mathcal{M}$ can be achieved as $(p, m^*)$. For each patch {$p_{h,w} \in \mathcal{P}_{x_i}$}, the patch-level anomaly score is calculated by 
\begin{equation}
s_{h,w}=\mathcal{W}_{m^*}\|p_{h,w} - m^*\|_2 , 
\end{equation}
where $\mathcal{W}_{m^*}$ is the soft weight calculated by one single discriminator. 
The image-level anomaly score is attained by finding the largest soft weights re-weighted patch-level anomaly score: 
$s^*=\mathop{\max}\limits_{(h,w)}s_{h,w}$.
 
{Specifically, SoftPatch only utilizes a single discriminator for coreset construction, and SoftPatch+  uses multiple single discriminators.}
Different from PatchCore which directly considers patches equally, SoftPatch and SoftPatch+ soften anomaly scores by noise level from noise discriminater. 
The soft weights, i.e., local outlier factors, have considered the local relationship around the nearest node. Thus, a similar effect can be achieved as PatchCore but with more noise robustness and fewer searches.
According to the image-level anomaly score, a sample is classified into a normal sample or an abnormal sample.

\section{{Fully Unsupervised Anomaly Detection Settings}}

{
Fully unsupervised ADs are more practical and challenging than common unsupervised ADs. In real-world industrial manufacturing, the ratio of corrupted products may be high (e.g., 38.5\%).
Meanwhile, existing unsupervised AD methods are susceptible to noisy data, especially in high noise ratios. The main reason is that they assume that the training dataset only contains normal samples. However, in practical applications, it isn't easy to guarantee that all training samples are normal. Hence, a more practical solution is to perform fully unsupervised AD, which allows a certain proportion of abnormal samples in the training process. This paradigm can better adapt to real-world scenarios and reduce reliance on manual annotation.}

{However, it isn't easy to create a fully unsupervised AD setting with independent training and testing sets when we use existing anomaly detection datasets, such as MVTec AD and VisA.} These benchmarks are usually divided into two sets including training and testing sets. The training set only contains normal samples, while the testing set consists of both normal and anomalous samples. To create a noisy training set, we have to sample anomalous samples randomly from the testing set, and then mix them with the normal training samples. In this way, it leads to a significant reduction in the number of testing samples, which may be insufficient to evaluate the performance of anomaly detection algorithms, especially for high noisy ratios.

{Notice that the number of original normal samples in the training set remains unchanged compared with the noiseless case. By controlling the proportion of noisy samples injected into the train set, we obtain several new datasets with different noise ratios dubbed noise-$n\%$, where $n\%$ refers to the ratio of noise. 
To comprehensively demonstrate the performance of fully unsupervised AD methods, we establish two different settings, no-overlap and overlap as follows.}

\begin{table*}[h]
\scriptsize
\setlength\tabcolsep{3pt}
\caption{\small{Fully unsupervised anomaly classification performance on MVTecAD with 10\% noise.}}
\label{tab-mvtec-cls}
\centering
\resizebox{1.0\textwidth}{!}{
    \begin{tabular}{c|c|cccc|c|ccc|c}
    \toprule
        \multirow{2}{*}{setting} & \multirow{2}{*}{Category} & \multirow{2}{*}{C-Flow} & \multirow{2}{*}{PaDiM} & \multirow{2}{*}{IGD} & \multirow{2}{*}{PatchCore} & \multirow{2}{*}{upperbound} & SoftPatch & SoftPatch & SoftPatch & \multirow{2}{*}{SoftPatch+} \\ 
          &  &  &  &  &  &  & Nearest & Gaussian & LOF &  \\ 
        \midrule
        \multirow{15}{*}{No Overlap}
        & bottle & 0.998 & 0.994 & 0.995 & 1.000 & 1.000 & 1.000 & 0.997 & 0.937 & 1.000 \\
        & cable & 0.925 & 0.873 & 0.826 & 0.982 & 0.996 & 0.935 & 0.952 & 0.995 & 0.990 \\
        & capsule & 0.947 & 0.920 & 0.868 & 0.976 & 0.978 & 0.916 & 0.662 & 0.963 & 0.971 \\
        & carpet & 0.961 & 0.999 & 0.817 & 0.996 & 0.989 & 0.995 & 0.999 & 0.991 & 0.940 \\
        & grid & 0.891 & 0.966 & 0.940 & 0.971 & 0.961 & 0.972 & 0.997 & 0.968 & 0.991 \\
        & hazelnut & 1.000 & 0.956 & 0.990 & 0.998 & 1.000 & 1.000 & 1.000 & 1.000 & 1.000 \\
        & leather & 1.000 & 1.000 & 0.868 & 1.000 & 1.000 & 1.000 & 1.000 & 1.000 & 1.000 \\
        & metal\_nut & 0.959 & 0.987 & 0.910 & 0.999 & 0.999 & 0.994 & 0.997 & 0.999 & 0.997 \\
        & pill & 0.929 & 0.918 & 0.786 & 0.975 & 0.974 & 0.921 & 0.873 & 0.963 & 1.000 \\
        & screw & 0.784 & 0.838 & 0.723 & 0.966 & 0.974 & 0.862 & 0.475 & 0.960 & 0.957 \\
        & tile & 0.991 & 0.977 & 0.971 & 0.985 & 0.984 & 0.996 & 0.997 & 0.993 & 0.995 \\
        & toothbrush & 0.906 & 0.927 & 1.000 & 0.997 & 0.997 & 1.000 & 0.997 & 0.997 & 1.000 \\
        & transistor & 0.896 & 0.953 & 0.893 & 0.953 & 1.000 & 1.000 & 0.992 & 0.990 & 0.995 \\
        & wood & 0.972 & 0.991 & 0.994 & 0.984 & 0.994 & 0.984 & 0.997 & 0.987 & 0.935 \\
        & zipper & 0.928 & 0.852 & 0.858 & 0.981 & 0.994 & 0.976 & 0.979 & 0.978 & 0.972 \\ \midrule
        & Average & 0.939 & 0.943 & 0.896 & 0.984 & 0.989 & 0.970 & 0.927 & 0.986 & 0.982 \\
        \midrule
        \midrule
        \multirow{15}{*}{Overlap} 
         & bottle  & 0.999  & 0.994  & 0.959  & 0.869  & 1.000  & 1.000  & 0.998  & 1.000  & 1.000  \\ 
         & cable  & 0.848  & 0.809  & 0.707  & 0.785  & 0.996  & 0.948  & 0.941  & 0.994  & 0.887  \\ 
         & capsule  & 0.767  & 0.802  & 0.554  & 0.865  & 0.980  & 0.928  & 0.729  & 0.943  & 0.984  \\ 
         & carpet  & 0.965  & 0.956  & 0.880  & 0.801  & 0.988  & 0.994  & 0.995  & 0.994  & 0.967  \\ 
         & grid  & 0.809  & 0.889  & 0.920  & 0.699  & 0.977  & 0.967  & 0.928  & 0.975  & 0.981  \\ 
         & hazelnut  & 0.990  & 0.714  & 0.880  & 0.406  & 1.000  & 1.000  & 0.997  & 0.998  & 0.991  \\ 
         & leather  & 0.993  & 0.981  & 0.791  & 0.771  & 1.000  & 1.000  & 1.000  & 1.000  & 1.000  \\ 
         & metal\_nut  & 0.978  & 0.926  & 0.787  & 0.840  & 0.999  & 0.994  & 0.989  & 0.999  & 0.999  \\ 
         & pill  & 0.927  & 0.808  & 0.845  & 0.767  & 0.972  & 0.909  & 0.881  & 0.951  & 1.000  \\ 
         & screw & 0.884  & 0.611  & 0.639  & 0.779  & 0.979  & 0.650  & 0.465  & 0.927  & 0.954  \\ 
         & tile  & 0.995  & 0.943  & 0.974  & 0.799  & 0.988  & 0.992  & 0.991  & 0.994  & 0.991  \\ 
         & toothbrush  & 0.942  & 0.847  & 0.969  & 0.939  & 0.997  & 1.000  & 1.000  & 0.986  & 0.995  \\ 
         & transistor  & 0.933  & 0.789  & 0.825  & 0.702  & 1.000  & 0.965  & 0.992  & 0.988  & 0.986  \\ 
         & wood  & 0.959  & 0.975  & 0.959  & 0.696  & 0.990  & 0.988  & 0.988  & 0.988  & 0.985  \\ 
         & zipper & 0.921  & 0.704  & 0.945  & 0.949  & 0.994  & 0.979  & 0.989  & 0.977  & 0.928  \\ \midrule
         & Average & 0.927  & 0.850  & 0.842  & 0.778  & 0.991  & 0.954  & 0.925  & 0.981  & 0.976\\
         \bottomrule
    \end{tabular}}
\end{table*}

\begin{table*}[h]
\scriptsize
\setlength\tabcolsep{3pt}
\caption{Fully unsupervised anomaly segmentation performance on MVTecAD with 10\% noise.}
\label{tab-mvtec-seg}
\centering
\resizebox{1.0\textwidth}{!}{
\begin{tabular}{c|c|cccc|c|ccc|c}
\toprule
  \multirow{2}{*}{setting} & \multirow{2}{*}{Category} & \multirow{2}{*}{C-Flow} & \multirow{2}{*}{PaDiM} & \multirow{2}{*}{IGD} & \multirow{2}{*}{PatchCore} & \multirow{2}{*}{upperbound} & SoftPatch & SoftPatch & SoftPatch & \multirow{2}{*}{SoftPatch+} \\ 
          &  &  &  &  &  &  & Nearest & Gaussian & LOF &  \\ 
\midrule
\multirow{15}{*}{No Overlap}          
         & bottle  & 0.984 & 0.986 & 0.767  & 0.987 & 0.988  & 0.987 & 0.986 & 0.987 & 0.987  \\ 
         & cable  & 0.958 & 0.916 & 0.725  & 0.843 & 0.986  & 0.915 & 0.981 & 0.983 & 0.984  \\ 
         & capsule  & 0.985 & 0.986 & 0.932  & 0.986 & 0.991  & 0.988 & 0.977 & 0.99 & 0.986  \\ 
         & carpet  & 0.989 & 0.992 & 0.951  & 0.992 & 0.991  & 0.992 & 0.993 & 0.992 & 0.986  \\ 
         & grid  & 0.947 & 0.974 & 0.965  & 0.991 & 0.989  & 0.99 & 0.989 & 0.99 & 0.991  \\ 
         & hazelnut  & 0.991 & 0.987 & 0.987  & 0.99 & 0.992  & 0.99 & 0.991 & 0.99 & 0.986  \\ 
         & leather  & 0.994 & 0.994 & 0.929  & 0.991 & 0.993  & 0.994 & 0.994 & 0.993 & 0.993  \\ 
         & metal\_nut  & 0.956 & 0.933 & 0.874  & 0.842 & 0.984  & 0.894 & 0.964 & 0.984 & 0.966  \\ 
         & pill  & 0.983 & 0.956 & 0.919  & 0.971 & 0.979  & 0.974 & 0.972 & 0.981 & 0.992  \\ 
         & screw & 0.977 & 0.989 & 0.977  & 0.995 & 0.995  & 0.991 & 0.969 & 0.994 & 0.971  \\ 
         & tile  & 0.953 & 0.956 & 0.794  & 0.953 & 0.967  & 0.96 & 0.962 & 0.954 & 0.965  \\ 
         & toothbrush  & 0.988 & 0.991 & 0.981  & 0.989 & 0.986  & 0.988 & 0.988 & 0.985 & 0.967  \\ 
         & transistor  & 0.887 & 0.96 & 0.891  & 0.847 & 0.975  & 0.965 & 0.954 & 0.942 & 0.969  \\ 
         & wood  & 0.964 & 0.973 & 0.909  & 0.969 & 0.969  & 0.947 & 0.946 & 0.939 & 0.984  \\ 
         & zipper & 0.978 & 0.986 & 0.795  & 0.986 & 0.990  & 0.989 & 0.988 & 0.988 & 0.988  \\
\midrule
         & Average & 0.969 & 0.972 & 0.893  & 0.956 & 0.985  & 0.971 & 0.977 & 0.979 & 0.981  \\ 
\midrule
\midrule
\multirow{15}{*}{Overlap} 
         & bottle  & 0.981  & 0.982  & 0.858  & 0.763  & 0.985  & 0.985  & 0.985  & 0.981  & 0.984  \\ 
         & cable  & 0.950  & 0.958  & 0.675  & 0.785  & 0.984  & 0.976  & 0.968  & 0.977  & 0.981  \\ 
         & capsule  & 0.980  & 0.977  & 0.934  & 0.798  & 0.990  & 0.984  & 0.980  & 0.986  & 0.982  \\ 
         & carpet  & 0.986  & 0.983  & 0.963  & 0.692  & 0.991  & 0.991  & 0.990  & 0.991  & 0.984  \\ 
         & grid  & 0.954  & 0.901  & 0.968  & 0.692  & 0.988  & 0.980  & 0.954  & 0.981  & 0.985  \\ 
         & hazelnut  & 0.979  & 0.978  & 0.978  & 0.750  & 0.987  & 0.985  & 0.971  & 0.967  & 0.991  \\ 
         & leather  & 0.991  & 0.994  & 0.929  & 0.791  & 0.993  & 0.993  & 0.993  & 0.993  & 0.983  \\ 
         & metal\_nut  & 0.939  & 0.946  & 0.887  & 0.739  & 0.983  & 0.897  & 0.953  & 0.947  & 0.993  \\ 
         & pill  & 0.969  & 0.956  & 0.931  & 0.919  & 0.978  & 0.974  & 0.969  & 0.976  & 0.986  \\ 
         & screw & 0.969  & 0.969  & 0.978  & 0.858  & 0.995  & 0.987  & 0.942  & 0.966  & 0.971  \\ 
         & tile  & 0.956  & 0.930  & 0.762  & 0.721  & 0.957  & 0.945  & 0.946  & 0.951  & 0.952  \\ 
         & toothbrush  & 0.986  & 0.984  & 0.961  & 0.960  & 0.986  & 0.986  & 0.986  & 0.983  & 0.949  \\ 
         & transistor  & 0.789  & 0.956  & 0.789  & 0.682  & 0.963  & 0.936  & 0.940  & 0.944  & 0.950  \\ 
         & wood  & 0.952  & 0.946  & 0.862  & 0.615  & 0.949  & 0.935  & 0.934  & 0.943  & 0.987  \\ 
         & zipper & 0.953  & 0.977  & 0.973  & 0.911  & 0.989  & 0.987  & 0.985  & 0.987  & 0.987  \\ 
\midrule
         & Average & 0.956  & 0.962  & 0.897  & 0.779  & 0.981  & 0.970  & 0.966  & 0.972  & 0.978  \\ 
\bottomrule
\end{tabular}}
\end{table*}

{ \emph{No Overlap:} 
 In this setting, we randomly selected anomalous samples from the testing set and {added} them to the normal training set. These added anomalous samples will be seen as noises since they may be taken as normal samples at the training stage. Meanwhile, these selected anomalous samples will be removed from the testing set, and we evaluate the model only on the remaining testing samples. Under the \emph{No Overlap} mode, it may not be possible to evaluate the model when all the anomalous samples in the testing set are selected as noises. This is true when using a higher noise ratio for the existing anomaly detection benchmarks where the proportion of abnormal data is usually relatively small. Therefore, we set the highest noise ratio as $10\%$ for MVTecAD and $8\%$ for VisA under the \emph{No Overlap} mode.}

{\emph{Overlap:} To evaluate the effectiveness of our method with high noisy ratios, we construct the \emph{Overlap} mode where those selected anomalous samples are still included in the testing set. This setting is more challenging and practical. There is a serious label conflict issue between the training and inference stages, where a sample may be taken as normal at the training phase but needs to be recognized anomaly at the testing stage, which makes this task challenging. Considering that in real industrial inspection scenarios, it is highly consistent for the appearance of the same products. In addition, there may be some slight changes in the imaging due to some factors such as lighting and camera shaking. In order to simulate this interference, we perform appropriate data augmentation on the sampled noisy images, such as rotation and Gaussian noise. {The data augmentation mainly includes rotation and translation transformations. Specifically, the rotation angle is within the range of [-3, 3] degrees. The translation includes vertical and horizontal directions. The number of translated pixels is a multiple of the height and width of the original image, and the range is between [-0.03, 0.03].}
In short, the \emph{Overlap} mode can ensure that we build noisy anomaly detection experiments with high noise ratios (such as 10\%-40\%), which is closer to real-world industrial inspection applications.}

\emph{Dicussion:} 
{The \emph{overlap} setting is both practical and reasonable for real-world production lines, where defects often appear similar across training and testing sets. This setting is simulated by applying data augmentation to anomaly testing images and incorporating them into the training set. Current anomaly detection datasets are limited in size and lack sufficient similar defects, making a no-overlap division less representative of real scenarios. Unlike traditional machine learning, where independent splits prevent overfitting, fully unsupervised anomaly detection does not face overfitting issues with overlap settings. Instead, it introduces challenges, as demonstrated by the performance decline of methods like PatchCore with increasing noise ratios in the overlap setting, as shown in Figure~\ref{chart1}.}

%% file: secs/4_experiments.tex
\section{Experiments} \label{section:exp}

\subsection{Experimental Details}
 \paragraph{Datasets} Our experiments are mainly conducted on the MVTecAD\cite{16}, BTAD\cite{btad} and {VisA}\cite{visa} benchmarks. MVTecAD contains 15 categories with 3629 training images and 1725 testing images in total, and BTAD has three categories with 1799 images, where different classes of industry production mean a comprehensive challenge, such as object or texture and whether rotation. VisA\cite{visa} is the largest industrial anomaly detection dataset. It contains 10821 high-resolution color images covering 12 objects in 3 domains.

\paragraph{Evaluation Metrics} We report both image-level and pixel-level AUROC for each category and average them to get the average image/pixel-level AUROC. In order to represent noise robustness, the performance gaps between noise-free data and noisy data are also displayed. When not otherwise stated, our method SoftPatch refers to SoftPatch-LOF that uses LOF in {Sec.} \ref{sec:lof}, and SoftPatch+ uses the fusion of Gaussian and LOF in {Sec.} \ref{sec:mdf}.

\begin{figure}[h]
  \centering
  \includegraphics[width=\linewidth]{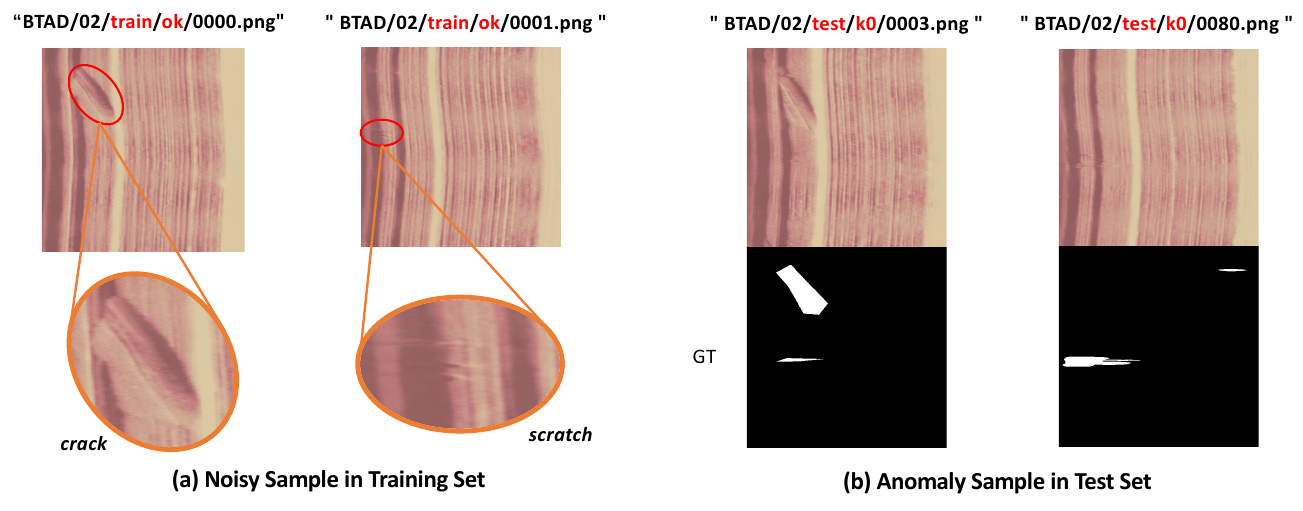}
  \caption{
  Noisy examples and corresponding anomaly samples in BTAD-02.}
  \label{fig:noise-samples-btad}
\end{figure}

\begin{figure*}[ht]
  \centering
    \includegraphics[width=1.0\linewidth]{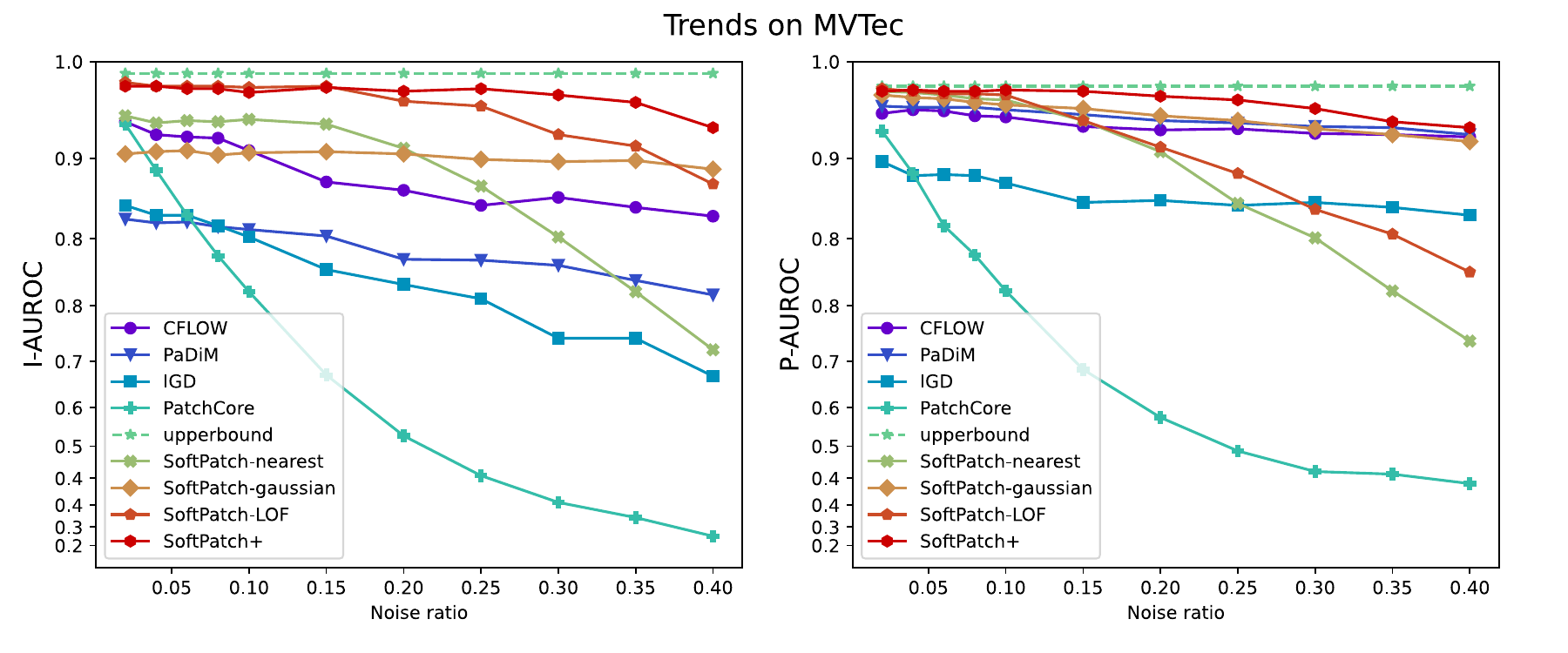}
    \includegraphics[width=1.0\linewidth]{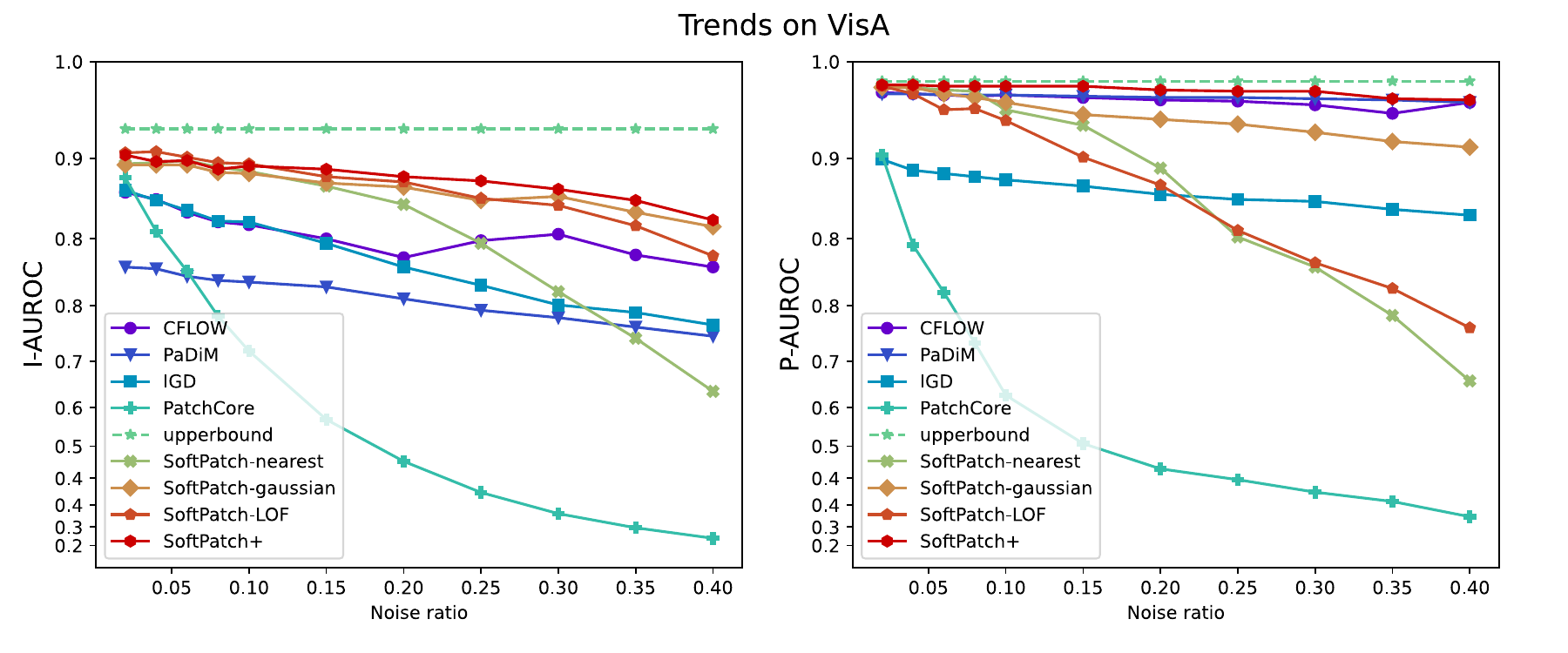}
  \vspace{-10pt}
  \caption{The comparison of performance trends {on MVTec and VisA} with the proposed SoftPatch and SoftPatch+, and the state-of-the-art methods with the noise ratio increases from 2\% to 40\%, {e.g., \{2\%, 5\%, 10\%, 15\%, 20\%, 25\%, 30\%, 35\%, 40\%\}}.}
  \label{chart1}
\end{figure*}

\begin{table*}[h]
\scriptsize
\setlength\tabcolsep{3pt}
\caption{Fully unsupervised anomaly classification performance on VisA with 8\% noise.}
\label{tab-VisA-cls}
\centering
\resizebox{1.0\textwidth}{!}{
\begin{tabular}{c|c|cccc|c|ccc|c}
\toprule
 \multirow{2}{*}{setting} & \multirow{2}{*}{Category} & \multirow{2}{*}{C-Flow} & \multirow{2}{*}{PaDiM} & \multirow{2}{*}{IGD} & \multirow{2}{*}{PatchCore} & \multirow{2}{*}{upperbound} & SoftPatch & SoftPatch & SoftPatch & \multirow{2}{*}{SoftPatch+} \\ 
          &  &  &  &  &  &  & Nearest & Gaussian & LOF &  \\ 
\midrule
\multirow{15}{*}{No Overlap} 
         & Candle & 0.930  & 0.868  & 0.834  & 0.990  & 0.991  & 0.970  & 0.944  & 0.975  & 0.734  \\ 
         & Capsules & 0.775  & 0.645  & 0.756  & 0.773  & 0.762  & 0.703  & 0.682  & 0.760  & 0.987  \\ 
         & Cashew & 0.920  & 0.908  & 0.913  & 0.987  & 0.989  & 0.989  & 0.985  & 0.990  & 0.948  \\ 
         & Chewinggum & 0.997  & 0.973  & 0.897  & 0.980  & 0.991  & 0.982  & 0.988  & 0.980  & 0.990  \\ 
         & Fryum & 0.946  & 0.862  & 0.914  & 0.965  & 0.954  & 0.946  & 0.940  & 0.945  & 0.956  \\ 
         & Macaroni1 & 0.887  & 0.758  & 0.857  & 0.948  & 0.975  & 0.954  & 0.846  & 0.948  & 0.985  \\ 
         & Macaroni2 & 0.844  & 0.639  & 0.557  & 0.748  & 0.767  & 0.759  & 0.717  & 0.738  & 0.948  \\ 
         & Pcb1 & 0.884  & 0.851  & 0.828  & 0.985  & 0.994  & 0.960  & 0.927  & 0.983  & 0.789  \\ 
         & Pcb2 & 0.820  & 0.746  & 0.701  & 0.975  & 0.979  & 0.911  & 0.890  & 0.956  & 0.985  \\ 
         & Pcb3 & 0.806  & 0.715  & 0.854  & 0.987  & 0.987  & 0.893  & 0.928  & 0.977  & 0.919  \\ 
         & Pcb4 & 0.973  & 0.952  & 0.989  & 0.994  & 1.000  & 1.000  & 0.991  & 0.996  & 0.961  \\ 
         & Pipe Fryum & 0.942  & 0.889  & 0.926  & 0.993  & 0.997  & 0.993  & 0.989  & 0.994  & 0.999  \\ \midrule
         & Average & 0.894  & 0.817  & 0.836  & 0.944  & 0.949  & 0.921  & 0.902  & 0.937  & 0.933  \\ \midrule
         \midrule
         \multirow{15}{*}{Overlap} 
         & Candle & 0.927  & 0.787  & 0.890  & 0.709  & 0.995  & 0.959  & 0.964  & 0.971  & 0.611  \\ 
         & Capsules & 0.709  & 0.565  & 0.668  & 0.540  & 0.764  & 0.651  & 0.630  & 0.646  & 0.968  \\ 
         & Cashew & 0.948  & 0.903  & 0.933  & 0.752  & 0.974  & 0.968  & 0.970  & 0.962  & 0.988  \\ 
         & Chewinggum & 0.991  & 0.977  & 0.869  & 0.794  & 0.988  & 0.979  & 0.987  & 0.991  & 0.947  \\ 
         & Fryum & 0.933  & 0.826  & 0.905  & 0.805  & 0.957  & 0.945  & 0.932  & 0.940  & 0.956  \\ 
         & Macaroni1 & 0.678  & 0.764  & 0.861  & 0.648  & 0.977  & 0.955  & 0.887  & 0.957  & 0.992  \\ 
         & Macaroni2 & 0.658  & 0.641  & 0.512  & 0.455  & 0.756  & 0.720  & 0.712  & 0.636  & 0.950  \\ 
         & Pcb1 & 0.887  & 0.861  & 0.917  & 0.857  & 0.986  & 0.950  & 0.951  & 0.974  & 0.658  \\ 
         & Pcb2 & 0.856  & 0.714  & 0.919  & 0.811  & 0.972  & 0.932  & 0.934  & 0.933  & 0.970  \\ 
         & Pcb3 & 0.830  & 0.647  & 0.916  & 0.834  & 0.988  & 0.893  & 0.932  & 0.970  & 0.934  \\ 
         & Pcb4 & 0.950  & 0.943  & 0.977  & 0.929  & 0.996  & 0.997  & 0.997  & 0.995  & 0.952  \\ 
         & Pipe Fryum & 0.930  & 0.876  & 0.924  & 0.818  & 0.998  & 0.992  & 0.990  & 0.992  & 0.998  \\ 
         \midrule
         & Average & 0.858  & 0.792  & 0.858  & 0.746  & 0.946  & 0.912  & 0.907  & 0.914  & 0.910 \\ 
            \bottomrule
\end{tabular}}
\end{table*}
\begin{table*}[h]
\scriptsize
\setlength\tabcolsep{3pt}
\caption{\small{Fully unsupervised anomaly segmentation performance on VisA with 8\% noise.}}
\label{tab-VisA-seg}
\centering
\resizebox{1.0\textwidth}{!}{
\begin{tabular}{c|c|cccc|c|ccc|c}
\toprule
 \multirow{2}{*}{setting} & \multirow{2}{*}{Category} & \multirow{2}{*}{C-Flow} & \multirow{2}{*}{PaDiM} & \multirow{2}{*}{IGD} & \multirow{2}{*}{PatchCore} & \multirow{2}{*}{upperbound} & SoftPatch & SoftPatch & SoftPatch & \multirow{2}{*}{SoftPatch+} \\ 
          &  &  &  &  &  &  & Nearest & Gaussian & LOF &  \\  
\midrule
\multirow{15}{*}{No Overlap} 
         & Candle & 0.990  & 0.986  & 0.908  & 0.995  & 0.994  & 0.991  & 0.990  & 0.995  & 0.985  \\ 
         & Capsules & 0.939  & 0.926  & 0.945  & 0.986  & 0.991  & 0.978  & 0.983  & 0.982  & 0.980  \\ 
         & Cashew & 0.983  & 0.981  & 0.664  & 0.985  & 0.985  & 0.983  & 0.985  & 0.983  & 0.995  \\ 
         & Chewinggum & 0.994  & 0.993  & 0.803  & 0.995  & 0.994  & 0.996  & 0.995  & 0.992  & 0.916  \\ 
         & Fryum & 0.959  & 0.960  & 0.878  & 0.926  & 0.925  & 0.923  & 0.921  & 0.922  & 0.994  \\ 
         & Macaroni1 & 0.994  & 0.990  & 0.992  & 0.999  & 0.999  & 0.999  & 0.998  & 0.999  & 0.989  \\ 
         & Macaroni2 & 0.965  & 0.948  & 0.971  & 0.979  & 0.981  & 0.986  & 0.983  & 0.979  & 0.999  \\ 
         & Pcb1 & 0.995  & 0.992  & 0.840  & 0.745  & 1.000  & 0.865  & 0.999  & 1.000  & 0.977  \\ 
         & Pcb2 & 0.943  & 0.978  & 0.930  & 0.992  & 0.993  & 0.989  & 0.983  & 0.992  & 0.999  \\ 
         & Pcb3 & 0.980  & 0.988  & 0.997  & 0.998  & 0.999  & 0.998  & 0.998  & 0.999  & 0.988  \\ 
         & Pcb4 & 0.983  & 0.978  & 0.970  & 0.977  & 0.992  & 0.987  & 0.988  & 0.987  & 0.998  \\ 
         & Pipe Fryum & 0.982  & 0.990  & 0.983  & 0.989  & 0.989  & 0.988  & 0.989  & 0.988  & 0.990  \\ 
\midrule
         & Average & 0.976  & 0.976  & 0.907  & 0.964  & 0.987  & 0.974  & 0.984  & 0.985  & 0.984  \\ 
\midrule
\midrule
\multirow{15}{*}{Overlap} 
         & Candle & 0.986  & 0.985  & 0.929  & 0.609  & 0.993  & 0.991  & 0.991  & 0.993  & 0.977  \\ 
         & Capsules & 0.968  & 0.926  & 0.951  & 0.683  & 0.993  & 0.952  & 0.928  & 0.942  & 0.985  \\ 
         & Cashew & 0.988  & 0.977  & 0.554  & 0.551  & 0.987  & 0.985  & 0.985  & 0.986  & 0.991  \\ 
         & Chewinggum & 0.992  & 0.989  & 0.791  & 0.596  & 0.989  & 0.992  & 0.990  & 0.938  & 0.910  \\ 
         & Fryum & 0.957  & 0.959  & 0.859  & 0.754  & 0.926  & 0.910  & 0.899  & 0.877  & 0.992  \\ 
         & Macaroni1 & 0.978  & 0.988  & 0.994  & 0.807  & 0.997  & 0.997  & 0.989  & 0.996  & 0.988  \\ 
         & Macaroni2 & 0.942  & 0.961  & 0.984  & 0.781  & 0.988  & 0.960  & 0.969  & 0.868  & 0.996  \\ 
         & Pcb1 & 0.987  & 0.988  & 0.878  & 0.613  & 0.998  & 0.997  & 0.995  & 0.998  & 0.979  \\ 
         & Pcb2 & 0.956  & 0.978  & 0.954  & 0.833  & 0.987  & 0.983  & 0.976  & 0.983  & 0.984  \\ 
         & Pcb3 & 0.972  & 0.978  & 0.977  & 0.737  & 0.994  & 0.993  & 0.992  & 0.990  & 0.998  \\ 
         & Pcb4 & 0.971  & 0.972  & 0.959  & 0.806  & 0.983  & 0.975  & 0.961  & 0.964  & 0.979  \\ 
         & Pipe Fryum & 0.982  & 0.992  & 0.974  & 0.730  & 0.989  & 0.989  & 0.990  & 0.988  & 0.992  \\ 
\midrule
         & Average & 0.973  & 0.974  & 0.900  & 0.708  & 0.985  & 0.977  & 0.972  & 0.960  & 0.981  \\ 
\bottomrule
\end{tabular}}
\end{table*}

\begin{figure*}[h]
  \centering
  \includegraphics[width=1.0\linewidth]{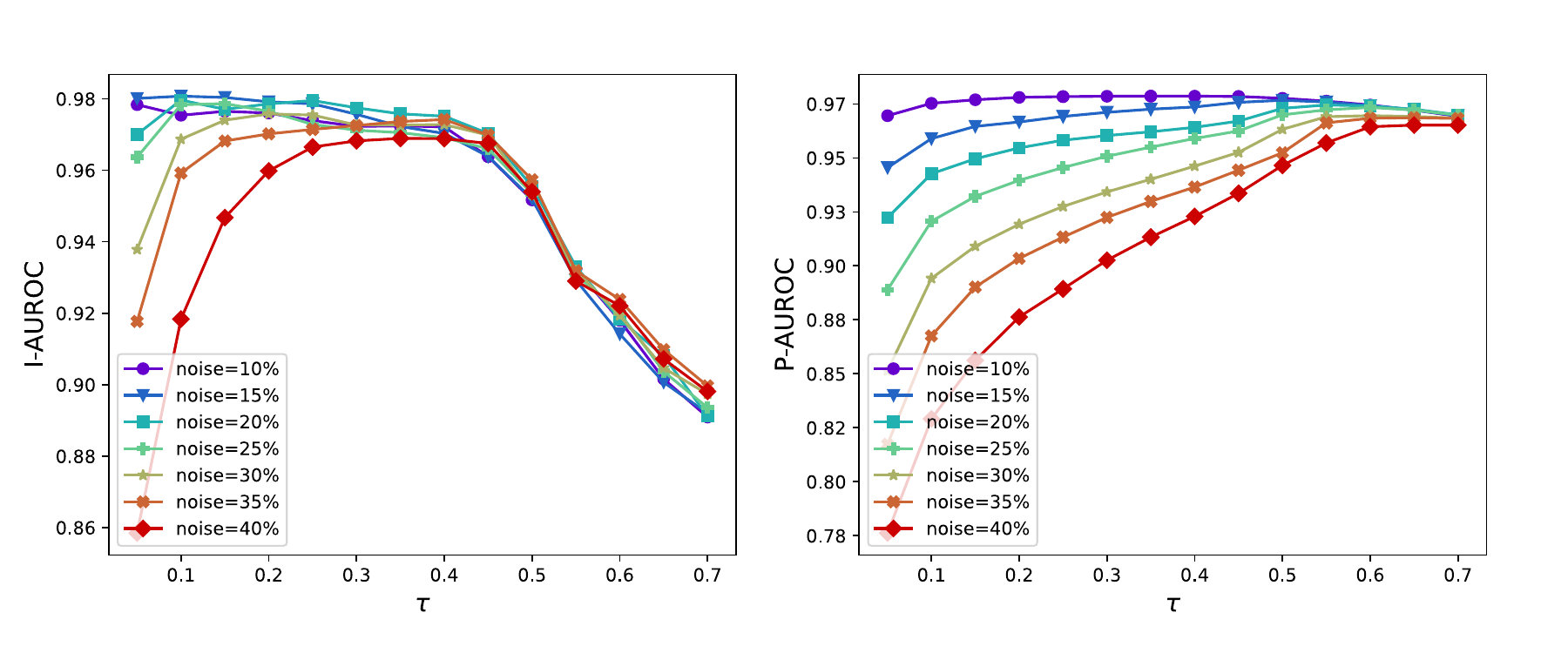}
   \vspace{-10pt}
  \caption{The performance trends of fully unsupervised anomaly classification and segmentation with the threshold $\tau$ increase (from 0.05 to 0.7) in SoftPatch+ when using different noisy ratios.}
  \label{fig-threshold}
\end{figure*}

\label{impl_details}
\paragraph{Implementation Details.} We test four SOTA AD algorithms, PatchCore~\citep{roth2021towards}, PaDim~\citep{defard2021padim} and CFLOW~\citep{gudovskiy2022cflow} and {IGD~\citep{chen2022deep}} in noise scene and follow their main settings. In the absence of specific instructions, the backbone of the feature extractor is \emph{Wide-ResNet50}, and the coreset sampling ratio of PatchCore and SoftPatch is 10\%. 
For MVTecAD images, we only use $256\times256$ resolution and center crops them into $224\times224$ along with a normalization. 
For BTAD, we use $512\times512$ resolution. 
We train a separate model for each class. 
Notice that unlike many methods setting the hyperparameters according to the noise ratio, which is unknowable in reality, we set the threshold $\tau$ in SoftPatch and the \emph{LOF-K} to constant 0.15 and 6 for all noisy scenarios and classes. 
{For SoftPatch+, we use two coresets to evaluate image-level classification ($\emph{LOF-K}=6, \tau_{cls}=0.15$) and pixel-level segmentation ($\emph{LOF-K}=6, \tau_{seg}=0.50$), respectively.}
The effects of hyperparameters are studied in the ablation study. All our experiments are run on Nvidia V100 GPU and repeated three times to report the average results.

{\subsection{Comparisons with State-of-the-Arts for Fully Unsupervised Industrial AD}}
\textbf{Robustness with Varying Noisy Ratios.}  
{In order to explore how different methods behave with the increasing noise level, experiments are performed on MVTecAD and VisA with different noisy ratios {\{2\%, 5\%, 10\%, 15\%, 20\%, 25\%, 30\%, 35\%, 40\%\}, covering the range from 2\% to 40\%} under the overlap setting. The results of the proposed methods (SoftPatch and SoftPatch+) and state-of-the-art (PaDim, C-Flow, and PatchCore) are provided in {Figure} \ref{chart1}.
First, as the noise ratio increases, PatchCore shows catastrophic drops both in image-level AUROC (about 70\%) and pixel-level AUROC (about 50\%). This means that the PatchCore is very sensitive to the quality of the features stored in the memory bank, which can be significantly affected by the presence of noise in training data. Second, we can see that distribution-based methods (IGD, PaDim, and C-Flow) are more robust than memory-bank-based PatchCore. However, there are larger performance gaps between these state-of-the-art methods and our upperbound (PatchCore using clear normal training images). Last but not least, our SoftPach+ achieves the most robust performance when using different noise ratios and different evaluation metrics. For SoftPatch, the nearest discriminator is the weakest among the three ones.
SoftPatch with LOF discriminator is better than  SoftPatch with Gaussian discriminator in image-level AUROC, but the results are opposite in pixel-level AUROC. Our SoftPatch+ integrates the advantages of LOF and Gaussian discriminators and achieves the best performance on two metrics.
} 

\textbf{Experiments on MVTecAD.} As indicated in Table \ref{tab-mvtec-cls} and Table \ref{tab-mvtec-seg}, when 10\% of anomalous samples are added to corrupt the train set, all existing methods have different extents of performance decrease, although not disastrously in \emph{No Overlap} setting. Compared to other methods, the proposed SoftPatch exhibits much stronger robustness against noisy data both in terms of anomaly classification and {segmentation}, no matter which single discriminator is used. Among {the} three variants of SoftPatch, SoftPatch-LOF achieves the best overall performance with the highest accuracy and strongest robustness. Interestingly, PaDiM\cite{defard2021padim}, CFLOW\cite{gudovskiy2022cflow}, and SoftPatch-gaussian show significantly less performance drop than PatchCore, which indicates that modeling feature as Gaussian distribution does help denoising. While modeling feature distribution at each spatial location as a single Gaussian distribution can't handle misaligned images, such as \textit{screw} class in MVTecAD, which explains the poor performance on these classes(see screw row). On the other hand, PatchCore's {greedy sampling} strategy is a double-edged sword with higher feature space coverage and higher sensitivity to noise. SoftPatch-nearest does a slightly better job in the misaligned cases. However, it doesn't take feature distribution into account, which leads to inferior performance.

\begin{table*}[h]
\scriptsize
\setlength\tabcolsep{3pt}
\caption{\small{Anomaly classification performance on BTAD without additional noise.}}
\label{tab-btad-cls}
\centering
\resizebox{\textwidth}{!}{
\begin{tabular}{c|cccc|c|ccc|c}
\toprule
 \multirow{2}{*}{Category} & \multirow{2}{*}{C-Flow} & \multirow{2}{*}{PaDiM} & \multirow{2}{*}{IGD} & \multirow{2}{*}{PatchCore} & \multirow{2}{*}{upperbound} & SoftPatch & SoftPatch & SoftPatch & \multirow{2}{*}{SoftPatch+} \\ 
            &  &  &  &  &  & Nearest & Gaussian & LOF &   \\   
\midrule
        01 & 0.992  & 1.000  & 0.936  & 1.000  & 1.000  & 0.995  & 0.996  & 0.999  & 0.999   \\ 
        02 & 0.996  & 0.871  & 0.761  & 0.871  & 0.867  & 0.873  & 0.880  & 0.934  & 0.910   \\ 
        03 & 0.869  & 0.971  & 0.999  & 0.999  & 0.999  & 0.997  & 0.987  & 0.997  & 0.995   \\ 
\midrule
        Average & 0.952  & 0.947  & 0.899  & 0.957  & 0.955  & 0.955  & 0.954  & 0.977  & 0.968   \\ 
\bottomrule
\end{tabular}}
\end{table*}
\begin{table*}[htb]
\scriptsize
\setlength\tabcolsep{3pt}
\caption{\small{Anomaly segmentation performance on BTAD without additional noise.}}
\label{tab-btad-seg}
\centering
\resizebox{\textwidth}{!}{
\begin{tabular}{c|cccc|c|ccc|c}
\toprule
 \multirow{2}{*}{Category} & \multirow{2}{*}{C-Flow} & \multirow{2}{*}{PaDiM} & \multirow{2}{*}{IGD} & \multirow{2}{*}{PatchCore} & \multirow{2}{*}{upperbound} & SoftPatch & SoftPatch & SoftPatch & \multirow{2}{*}{SoftPatch+} \\ 
            &  &  &  &  &  & Nearest & Gaussian & LOF &   \\  
\midrule
        1 & 0.980  & 0.978  & 0.894  & 0.977  & 0.977  & 0.976  & 0.971  & 0.976  & 0.971   \\ 
        2 & 0.990  & 0.991  & 0.930  & 0.968  & 0.968  & 0.964  & 0.965  & 0.966  & 0.968   \\ 
        3 & 0.961  & 0.960  & 0.889  & 0.993  & 0.993  & 0.991  & 0.987  & 0.994  & 0.992   \\ 
\midrule
        Average & 0.977  & 0.977  & 0.904  & 0.979  & 0.979  & 0.977  & 0.974  & 0.979  & 0.977  \\ 
\bottomrule
\end{tabular}}
\end{table*}
\begin{table*}[h]
\centering
\caption{{Effects of single-discriminator and multi-discriminator with varying noisy ratios (from 10\% to 40\%) on MVTecAD. The default modules and the corresponding results are highlighted in light gray background.}}
\label{tab:soft-weight}
\small
\resizebox{1.0\textwidth}{!}{
\begin{tabular}{ccc|cc|cc|cc|cc}
\toprule
  \multirow{2}{*}{nearest} &\multirow{2}{*}{gaussian} &\multirow{2}{*}{LOF}   &\multicolumn{2}{c|}{ 10\%}   &\multicolumn{2}{c|}{ 20\%}           &\multicolumn{2}{c|}{ 30\%}  &\multicolumn{2}{c}{ 40\%}  \\
\cmidrule{4-11}
 & &  & I-AUROC & P-AUROC & I-AUROC & P-AUROC & I-AUROC & P-AUROC & I-AUROC & P-AUROC \\
\midrule

\cmark	&\xmark	&\xmark			&\pmerror{0.956}{0.001}		&\pmerror{0.970}{0.000}		&\pmerror{0.928}{0.002}		&\pmerror{0.928}{0.001}		&\pmerror{0.843}{0.003}		&\pmerror{0.840}{0.001}		&\pmerror{0.700}{0.005}		&\pmerror{0.710}{0.002}	\\
																
\xmark	&\cmark	&\xmark			&\pmerror{0.926}{0.003}		&\pmerror{0.965}{0.001}		&\pmerror{0.923}{0.003}		&\pmerror{0.956}{0.001}		&\pmerror{0.918}{0.001}		&\pmerror{0.944}{0.002}		&\pmerror{0.911}{0.002}		&\pmerror{0.933}{0.002}	\\
															
\rowcolor[HTML]{D3D3D3} {\xmark}	&{\xmark}	&{\cmark}			&\pmerror{0.981}{0.003}		&\pmerror{0.974}{0.001}		&\pmerror{0.971}{0.003}		&\pmerror{0.932}{0.001}		&\pmerror{0.945}{0.001}		&\pmerror{0.876}{0.002}		&\pmerror{0.902}{0.002}		&\pmerror{0.809}{0.002}	\\
\midrule
\cmark	&\cmark	&\xmark			&\pmerror{0.925}{0.002}		&\pmerror{0.977}{0.000}		&\pmerror{0.911}{0.002}		&\pmerror{0.973}{0.000}		&\pmerror{0.878}{0.001}		&\pmerror{0.963}{0.002}		&\pmerror{0.808}{0.004}		&\pmerror{0.945}{0.001}	\\
\cmark	&\xmark	&\cmark			&\pmerror{0.979}{0.001}		&\pmerror{0.972}{0.000}		&\pmerror{0.974}{0.001}		&\pmerror{0.965}{0.000}		&\pmerror{0.933}{0.003}		&\pmerror{0.954}{0.001}		&\pmerror{0.836}{0.004}		&\pmerror{0.937}{0.001}	\\
\rowcolor[HTML]{D3D3D3} {\xmark}	&{\cmark}	&{\cmark}			&\pmerror{0.977}{0.002}		&\pmerror{0.951}{0.000}		&\pmerror{0.978}{0.002}		&\pmerror{0.950}{0.000}		&\pmerror{0.973}{0.005}		&\pmerror{0.945}{0.001}		&\pmerror{0.950}{0.006}		&\pmerror{0.931}{0.001}	\\
\cmark	&\cmark	&\cmark			&\pmerror{0.975}{0.002}		&\pmerror{0.977}{0.000}		&\pmerror{0.977}{0.002}		&\pmerror{0.961}{0.000}		&\pmerror{0.952}{0.002}		&\pmerror{0.940}{0.001}		&\pmerror{0.886}{0.006}		&\pmerror{0.900}{0.002}		\\
\bottomrule
\end{tabular}
}
\end{table*}

{\textbf{Experiments on VisA.} We report the anomaly classification and segmentation results of all SOTA and our SoftPatch and SoftPatch+ on the largest industrial anomaly detection benchmark (VisA) in Tables.~\ref{tab-VisA-cls} and \ref{tab-VisA-seg}. We can obtain similar experimental conclusions as MVTecAD. First, our SoftPach and SoftPatch+ significantly outperform all SOTA methods (PaDiM\cite{defard2021padim}, CFLOW\cite{gudovskiy2022cflow}, IGD\citep{chen2022deep}, and PatchCore\citep{roth2021towards}) on both anomaly classification and segmentation tasks. Second, the performance of our method is more robust than the existing SOTA methods in both \emph{Overlap} and \emph{No Overlap} settings, while the SOTA methods either perform poorly or are not robust. For example, although the PatchCore\citep{roth2021towards} achieves excellent performance (0.944 in classification AUROC and 0.964 in segmentation AUROC ) under a \emph{No Overlap} setting, the corresponding performance dramatically drops (0.746 in classification AUROC and 0.708 in segmentation AUROC) when applied to a more changeling setting, i.e., \emph{Overlap}; The PaDim\cite{defard2021padim} and IGD ~\citep{chen2022deep} methods are also robust, but their performance is worse than ours.}

\textbf{Experiments on BTAD.}  
We also compare SoftPatch with other SOTA methods on another dataset, BTAD. Surprisingly, SoftPatch gives out a new SOTA result, even in the original setting that contains no additional noise (Table~\ref{tab-btad-cls} and \ref{tab-btad-seg}). By reviewing all the training samples, we find that there are already many noisy samples (usually small scratches) in the training set of category BTAD-02, which is more consistent with our setting and further demonstrates the necessity of our approach. {The number of noisy samples of each class is reported in Table~\ref{tab-btad-cls} and some noisy images are shown in Figure~\ref{fig:noise-samples-btad}. We can see that the BTAD-02 contains more anomaly samples with similar appearance anomalies. Our method attains significant improvement on the BTAD-02 class compared to others.} 

\subsection{Ablation Study}
\subsubsection{Effectiveness of the Proposed Modules}
We validated the effectiveness of proposed \textbf{single-discriminator} and \textbf{multi-discriminator}. 
{In Table~\ref{tab:mdf}, it can be seen that SoftPatch+ introducing multiple discriminators significantly enhances the robustness of the model compared to PatchCore and SoftPatch, especially under a high noise setting.}
As shown in Table \ref{tab:soft-weight}, among three decision choices of a single discriminator, LOF achieved the best balance between robustness and capacity, resulting in the most performance boost under all settings. {We further compare anomaly detection performance with SoftPach+ using multiple discriminators. We can see that the performance is the worst when combining nearest and Gaussian, and the other three combinations achieve almost identical performance on the MVTecAD dataset when setting the noise ratio to 10\%. What is more, the performances of these three methods vary greatly under high noisy ratios (e.g., 40\%). Considering the balance of effectiveness and efficiency, our best suggestion is to choose Gaussian and LOF as the two basic discriminators in SoftPatch+.} 

\subsubsection{Parameter Selection}
To explore the impact of two parameters (\emph{LOF-K} and threshold $\tau$) on the final performance, we perform parameters searching on our method. As in Table \ref{tab:LOF-K}, our method achieves better performance when \emph{LOF-K} is greater than 5, which suggests that our method is not sensitive to \emph{LOF-K}, as long as it is not too small or too large. If \emph{LOF-K} is too small, it fails to estimate the local density accurately because too few neighbors are considered. On the contrary, a large \emph{LOF-K} may lead to undesirable {cross-cluster connections} that can not capture real data distribution.

Threshold $\tau$ refers to the ratio of eliminated patch features when building coreset. {Figure} \ref{fig-threshold} indicates an increasing trend of AUROC as threshold $\tau$ increases in pixel-level AUROC, which is expected since a higher threshold means a more aggressive denoising strategy for anomaly segmentation. Note that the mistakenly sampled features are the direct reason for the drastic performance drop. Therefore more aggressive denoising improves the result significantly. {On the contrary, a larger ($\tau>0.35$) or smaller ($\tau<0.10$) threshold usually results in performance drops for image-level anomaly classification. This is reasonable because a larger threshold will significantly reduce the number of training samples, and a smaller threshold will make the training samples contain more noise. Considering the different sensibility of $\tau$ in anomaly classification and segmentation, we set specific values (i.e., 0.15 for classification and 0.50 for segmentation) for them in our SoftPatch+.}

\subsubsection{Effects of Image-Level and Patch-Level Denoising}
{
In order to demonstrate the effects of patch-level denoising manner, we compare three different denoising manners including no denoising, image-level denoising, and patch-level denoising. Here, no denoising means that all noise samples will be taken as normal during the training phase. The image- and patch-level denoising indicates some noisy samples need to be removed using different denoising granularities before modeling. The former means that if an image is diagnosed as abnormal, it will be removed from the training set, and the latter means that only when the patch-level features in the noise image are diagnosed as abnormal and are removed, and the remaining features will be retained. We conduct experiments on MVTecAD with different noisy ratios (from 10\% to 40\%) and the corresponding results are shown in Table~\ref{tab:denoisemanner}.}

{We can see that anomaly classification and segmentation performance is better when using image- or patch-level denoising compared to no denoising manner. This suggests that denoising takes an important role in unsupervised anomaly detection with noisy data. Furthermore, patch-level denoising is more effective than image-level denoising for SoftPatch and SoftPatch+. This is not surprising, because there is a fact that the number of true anomaly features is only a small percentage of all features in a noisy image from the patch-level perspective.
Therefore, patch-level denoising can retain normal features when removing true anomalous features, but image-level denoising does not.}

\subsubsection{Qualitative Results}

\begin{figure*}[h]
\includegraphics[width=\linewidth]{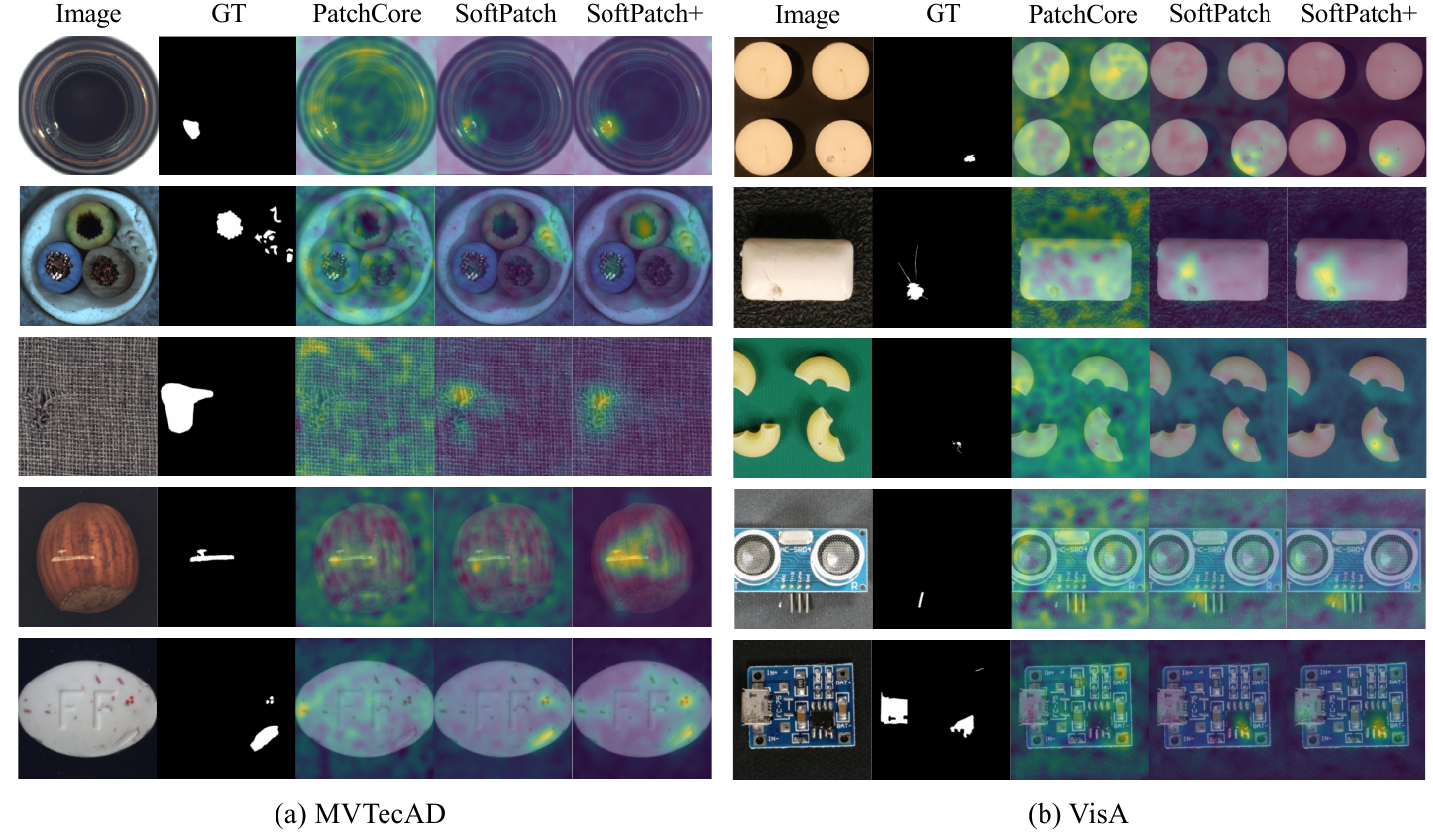}
\caption{{Visualizations of anomaly segmentation with PatchCore, SoftPatch, and SoftPatch+ on MVTecAD and VisA. These results demonstrate that the proposed method (SoftPatch and SoftPatch+) is more effective in localizing abnormal areas in both two datasets with a high noise rate (40\%).}} \label{fig:vis-res}
\end{figure*}

{Figure~\ref{fig:vis-res} shows results of anomaly
segmentation that indicates the abnormal areas on MVTecAD and VisA using a high noise rate (40\%). The experimental results demonstrate that the proposed SoftPach+ is more effective in localizing abnormal areas in both two datasets, even in challenging cases where the anomalies are small or difficult to detect. The method achieves high accuracy and precision in detecting anomalies. Furthermore, we found that the proposed method shows consistent performance across both object and texture classes, indicating its versatility and applicability to a wide range of anomaly detection tasks. Overall, the experimental results demonstrate that the proposed method is effective and robust for anomaly detection in images and has the potential to be applied in various industrial settings.}

\subsubsection{Computational Analysis}

SoftPatch does not require more runtime than PatchCore, according to theoretical analysis. The complexity of the greedy sampling process in PatchCore is $\mathcal{O}(N^2h^2w^2)$, which is the most expensive part. The complexity of the noise discrimination process in SoftPatch-LOF is $\mathcal{O}(N^2hw)$ since features are grouped before. So the computational complexity of SoftPatch is equal {to} PatchCore by $\mathcal{O}(N^2hw+N^2h^2w^2)=\mathcal{O}(N^2h^2w^2)$. In fact, SoftPatch will be faster because it removes a part of the patch as noise. {Furthermore, the complexity of our SoftPach+ should be $\mathcal{O}(N^3h^2w^2)$, which is from the multi-variate Gaussian distribution ($\mathcal{O}(N^3hw)$) and PatchCore ($\mathcal{O}(N^2h^2w^2)$). Although softpatch+ exhibits higher computational complexity, it brings robust anomaly detection performance when faced with high noise ratios. In the subsequent actual inference speed comparisons, we will see that the increasing time is acceptable, especially when chasing more robust and higher accuracy for real-world industrial scenarios.} 

{In order to further test the actual inference time of our SoftPatch and SoftPatch+. We conduct these experiments on a V100 GPU equipped with 32G Memory. Our SoftPatch is comparable to existing methods, with a high frame rate of 17 FPS. Furthermore, our SoftPatch+ is slightly slower (12 FPS) than SoftPatch, but it is more robust, especially in high-noise scenarios.}

\begin{table*}[t]
\centering
\scriptsize
\caption{Image- and pixel-level AUROC for different LOF-K on MVTecAD with 10\% noise.}
\label{tab:LOF-K}
\resizebox{1.0\textwidth}{!}{
\begin{tabular}{c|ccccccc}
\toprule
K          & 3           & 4           & 5           & 6           & 7           & 8           & 9           \\
\midrule
I-/P- AUROC    & \textbf{0.983}/0.955 & 0.982/0.951 & \textbf{0.983}/0.959 & 0.982/\textbf{0.975} & 0.981/0.973 & 0.982/0.968 & 0.980/0.968 \\
\bottomrule
\end{tabular}}
\end{table*}
\begin{table*}[h]
\centering
\caption{Effects of denoising manner on MVTecAD with different noisy ratios.}
\label{tab:denoisemanner}
\small
\resizebox{1.0\linewidth}{!}{
\begin{tabular}{c|c|cc|cc|cc|cc}
\toprule
\multirow{2}{*}{Methods} &\multirow{2}{*}{denoising}  &\multicolumn{2}{c|}{10\%}      &\multicolumn{2}{c|}{20\%}  &\multicolumn{2}{c|}{30\%}  &\multicolumn{2}{c}{40\%}       \\
\cmidrule{3-10}
  &  & I-AUROC & P-AUROC & I-AUROC & P-AUROC & I-AUROC & P-AUROC & I-AUROC & P-AUROC \\
\midrule
SoftPatch  &no &0.778 	&0.779 	&0.543 	&0.581 	&0.368 	&0.458 	&0.242 	&0.425    \\
SoftPatch  &image-level &0.935 	&0.889 	&0.751 	&0.807 	&0.586 	&0.675 	&0.448 	&0.586  \\
SoftPatch  &patch-level &0.980 	&0.974 	&0.969 	&0.930 	&0.941 	&0.871 	&0.896 	&0.802    \\
SoftPatch+ &patch-level &0.976 	&0.978 	&0.977 	&0.973 	&0.974 	&0.963 	&0.947 	&0.947  \\
\bottomrule
\end{tabular}
}
\end{table*}
\begin{table*}[h]
\centering
\caption{{Effects of single-discriminator and multi-discriminator with varying noisy ratios (from 10\% to 40\%) on MVTecAD. The best method is highlighted in light gray background.}}
\label{tab:mdf}
\small
\resizebox{1.0\textwidth}{!}{
\begin{tabular}{ccc|cc|cc|cc|cc}
\toprule
  \multirow{2}{*}{Methods} &\multirow{2}{*}{Single-Dis.} &\multirow{2}{*}{Multi-Dis.}  &\multicolumn{2}{c|}{ 10\%}   &\multicolumn{2}{c|}{ 20\%}           &\multicolumn{2}{c|}{ 30\%}  &\multicolumn{2}{c}{ 40\%}  \\
\cmidrule{4-11}
 & &  & I-AUROC & P-AUROC & I-AUROC & P-AUROC & I-AUROC & P-AUROC & I-AUROC & P-AUROC \\
\midrule
PatchCore		&\xmark	&\xmark		&\pmerror{0.773}{0.006}		&\pmerror{0.778}{0.002}		&\pmerror{0.546}{0.006}		&\pmerror{0.582}{0.002}		&\pmerror{0.366}{0.005}		&\pmerror{0.460}{0.001}		&\pmerror{0.239}{0.003}		&\pmerror{0.422}{0.002}	\\
SoftPatch		&\cmark	&\xmark		&\pmerror{0.981}{0.003}		&\pmerror{0.974}{0.001}		&\pmerror{0.971}{0.003}		&\pmerror{0.932}{0.001}		&\pmerror{0.945}{0.001}		&\pmerror{0.876}{0.002}		&\pmerror{0.902}{0.002}		&\pmerror{0.809}{0.002}	\\
\rowcolor[HTML]{D3D3D3} SoftPatch+		&\cmark	&\cmark		&\pmerror{0.977}{0.002}		&\pmerror{0.951}{0.000}		&\pmerror{0.978}{0.002}		&\pmerror{0.950}{0.000}		&\pmerror{0.973}{0.005}		&\pmerror{0.945}{0.001}		&\pmerror{0.950}{0.006}		&\pmerror{0.931}{0.001}	\\
\bottomrule
\end{tabular}
}
\end{table*}

%% file: secs/5_conclusion.tex
\section{Conclusion and Future Works} \label{section:con}
This paper emphasizes the practical value of investigating noisy data problems in unsupervised AD. Introducing a novel noisy setting on the previous task, we test the performance of existing methods and the proposed SoftPatch and SoftPatch+. For existing methods, despite no adaptation to noisy settings, some of them have a slight performance decrease in some scenes. However, the performance decrease could be more significant and catastrophic for other methods or in other scenes. The proposed SoftPatch and SoftPatch+ show consistent performance in all noise settings, which outperforms other methods. 

{
\textbf{Limitations.}
Although SoftPatch and SoftPatch+ perform well in various noisy environments, their performance may degrade when handling extremely high noise levels (exceeding 40\%). Additionally, the current methods are primarily designed for static image data, and their effectiveness on video or dynamic data has not yet been validated. Furthermore, our approach is developed based on PatchCore, resulting in high computational complexity and memory requirements, which may limit its applicability in resource-constrained environments.} {In addition, the current evaluation method may be related to the specific dataset distribution. More generalized conclusions need to be further verified under real-world noise distribution from production lines.}

{
\textbf{Future Works.}
Exploring more efficient algorithms to reduce computational complexity and memory requirements, beyond embedding-based PatchCore, could enable the application of these methods in resource-limited settings. Additionally, extending SoftPatch and SoftPatch+ to anomaly detection in video data or other dynamic data could validate their effectiveness in a broader range of applications. Moreover, investigating more diverse types and distributions of noise could further enhance the robustness and generalization capabilities of the methods. Finally, integrating semi-supervised or weakly supervised learning approaches, leveraging a small amount of labeled data, could further improve the performance of anomaly detection. We are also interested in how to construct a large-scale real-world anomaly detection dataset, simulating batch defects in actual production lines. In this way, independent training and testing settings can be established to promote research in the field of fully unsupervised anomaly detection.}

\noindent
\textbf{Acknowledgements}

This work was supported by the National Natural Science Foundation of China (Grant No. 62472282, Grant No. 72192821).